\documentclass[letterpaper,twocolumn,10pt]{article}

\usepackage{hegroup}
\usepackage[numbers,sort]{natbib}
\usepackage{tikz}
\usepackage{times}
\usepackage{helvet}
\usepackage{courier}
\usepackage{caption}
\usepackage{xspace}
\usepackage{amssymb}
\usepackage{makecell}
\usepackage{textcomp}
\usepackage{algorithm}
\usepackage{algorithmic}
\usepackage{booktabs}
\usepackage{multirow}
\usepackage{mathtools}
\usepackage{amsfonts}
\usepackage{bbm}
\usepackage{amsmath}
\usepackage{graphicx}
\usepackage{hyperref}
\usepackage{cleveref}
\usepackage{newfloat}
\usepackage{listings}
\usepackage{xcolor}
\usepackage{colortbl}
\usepackage{url}
\usepackage{subcaption}
\usepackage{bm} 
\newcommand{\mypara}[1]{\smallskip\noindent{\bf {#1}.}\xspace}
\newcommand{\Method}{BadVSFM\xspace}

\begin{document}

\title{Backdoor Attacks on Prompt-Driven Video Segmentation Foundation Models}

\author{
Zongmin Zhang\textsuperscript{1}\thanks{Equal contribution.}  \quad
Zhen Sun\textsuperscript{1}\footnotemark[1]  \quad 
Yifan Liao\textsuperscript{1}  \quad
Wenhan Dong\textsuperscript{1}  \\
Xinlei He\textsuperscript{1}\thanks{Corresponding author (\href{mailto:xinleihe@hkust-gz.edu.cn}{xinleihe@hkust-gz.edu.cn}).} \quad
Xingshuo Han\textsuperscript{2}  \quad 
Shengmin Xu\textsuperscript{3}  \quad 
Xinyi Huang\textsuperscript{4}  \quad
\\
\\
\textsuperscript{1}\textit{Hong Kong University of Science and Technology (Guangzhou)} \\
\textsuperscript{2}\textit{Nanjing University of Aeronautics and Astronautics} \\
\textsuperscript{3}\textit{Fujian Normal University} \quad 
\textsuperscript{4}\textit{Jinan University} \quad
\\
}
\date{}

\maketitle
%----------------------------------
\begin{abstract}
%----------------------------------

Prompt-driven Video Segmentation Foundation Models (VSFMs), such as SAM2, have been increasingly adopted in applications including autonomous driving and digital pathology, yet their security risks remain largely unexplored.
In this paper, we study backdoor attacks against VSFMs and show that directly applying classic attacks such as BadNet is largely ineffective, yielding attack success rates (ASR) below 5\%.
To understand this failure, we conduct gradient-similarity and attention-map analyses and find that traditional backdoor training fails to overcome the prompt-driven architecture of VSFMs: clean and triggered samples induce largely aligned image-encoder gradients, and model attention remains focused on the prompt-specified object rather than the trigger.
To address this limitation, we propose \Method, the first backdoor attack framework tailored to prompt-driven VSFMs.
\Method uses a two-stage training strategy that explicitly targets the two failure points above.
First, it steers triggered frames toward a designated target embedding while preserving clean-frame representations, forcing the encoder to learn trigger-specific features.
Second, it trains the decoder to map triggered frame prompt representations to an attacker-specified target mask while maintaining clean segmentation behavior, making the attack robust to different prompt types.
Extensive experiments on five VSFMs and two datasets show that \Method achieves strong and controllable backdoor effects across different triggers and prompts while largely preserving clean performance.
Systematic ablations over training stages, losses, attack targets, trigger configurations, and poisoning rates validate the robustness and necessity of the two-stage design.
Finally, interpretability analyses show that \Method separates clean and triggered representation directions and shifts attention toward trigger regions, while five representative defenses remain largely ineffective.
Our results reveal a practical and underexplored vulnerability of current VSFMs to backdoor threats.

% ----------------------------------------------------
\end{abstract}
% ----------------------------------------------------

\begin{center}
    \includegraphics[width=\linewidth]{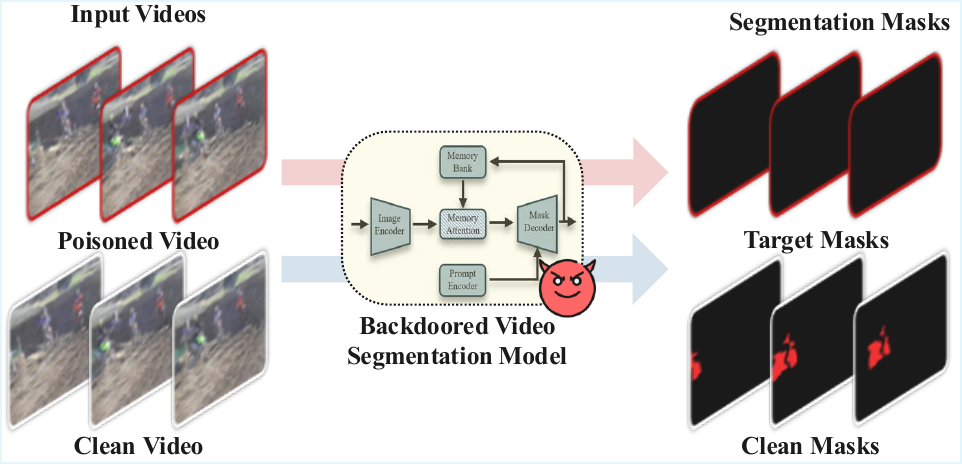}
    \captionof{figure}{Illustration of the effect of \Method on SAM2, comparing outputs under clean and triggered inputs.}
    \label{fig:overview}
\end{center}

% ----------------------------------------------------
\section{Introduction}
% ----------------------------------------------------

Video object segmentation is a fundamental task in computer vision that partitions video frames into foreground objects and background regions and produces object masks.
Recently, prompt-driven video segmentation foundation models (VSFMs) have rapidly advanced this task, with SAM2~\cite{DBLP:conf/iclr/RaviGHHR0KRRGMP25} emerging as a representative and influential example.
With strong generalization ability and efficient interactive segmentation, VSFMs are gradually becoming a mainstream choice in various applications~\cite{DBLP:journals/corr/abs-2503-12781}, including video retrieval~\cite{DBLP:journals/tist/YaoLXZZ20}, digital pathology~\cite{Zhang2024PathSAM2TS}, and medical video analysis~\cite{DBLP:journals/corr/abs-2408-00874,DBLP:journals/corr/abs-2408-03286}.
Despite being popular, the wide application of VSFMs also raises severe security concerns, with backdoor attacks as one of the critical threats~\cite{DBLP:journals/corr/abs-1708-06733,DBLP:conf/iccv/LiLWLHL21,DBLP:conf/sp/0001CLL0CH025,He2025AISecuritySurvey}.
By injecting samples with specific triggers during the training phase, the adversary embeds malicious behaviors into the models and later uses frames with the trigger during inference to induce incorrect predictions, thereby achieving the attack objective.
Such attacks can spread through various channels (e.g., releasing poisoned data or pretrained models on public platforms like Hugging Face), further expanding their impact.
Although backdoor attacks have been extensively studied in tasks such as image classification~\cite{DBLP:journals/corr/abs-1708-06733,DBLP:conf/iccv/LiLWLHL21,DBLP:journals/corr/abs-1712-05526}, their implications for VSFMs remain largely unexplored.
Attacking VSFMs is challenging because, unlike conventional vision models that map an input image directly to a label or mask, VSFMs generate segmentation masks conditioned jointly on video frames and user-provided prompts.
The additional prompt-processing module and prompt-conditioned decoding process may prevent existing backdoor mechanisms from reliably controlling the final mask prediction.
As an initial attempt, we apply a BadNet backdoor attack to VSFMs and observe that it is largely ineffective, achieving an attack success rate (ASR) of only around 5\% on both DAVIS~\cite{DBLP:journals/corr/Pont-TusetPCASG17} and LVOS~\cite{DBLP:conf/iccv/HongCLZGCZ23}.
This result suggests that directly transferring existing backdoor attacks to VSFMs fails to induce stable and controllable malicious behaviors.
To understand this phenomenon, we analyze encoder gradients and attention maps under clean and triggered frames and observe that, under traditional backdoor training, the encoder gradients for clean and triggered samples remain largely aligned, and the model’s attention still concentrates on the true object rather than the trigger region.
These results suggest that the image encoder and the prompt-driven decoder of VSFMs are jointly dominated by the clean segmentation objective: the encoder does not learn a distinct representation for the trigger, and the prompt supervision further steers the decoder to follow the ground-truth object instead of the attack target (more details can be found at~\Cref{sec:attack_performance}).
As a result, although traditional backdoor attacks remain effective on classification models, they fail to induce stable and controllable changes in the output masks of VSFMs, rendering a direct transfer of existing backdoor techniques largely ineffective.

To overcome this, we propose \Method, the first backdoor attack framework targeting VSFMs.
Contrarily, \Method applies a two-stage training strategy that explicitly drives the encoder to separate triggered inputs from clean inputs in the representation space, and then trains the decoder so that only the triggered representations are mapped to the backdoor target mask while the clean ones preserve the original segmentation behavior, as shown in~\Cref{fig:overview}.
Specifically, in stage 1, we fine-tune the image encoder so that triggered frames are mapped to a predefined target embedding, while clean frames are kept aligned with a clean reference encoder to preserve utility.
In stage 2, we update the mask decoder across different prompt types, so that the fusion of a triggered frame embedding with any prompt can be mapped to a shared target mask (e.g., an empty mask in our main experiment) while clean behavior is preserved via mask alignment with a reference decoder.
Concretely, we optimize the mask decoder with BCE and Dice losses, which we find most stable for maintaining both model utility and backdoor efficacy (See \Cref{sec:ablation_study} for more details).
By first aligning the representation space and then conditioning the decoder to a prompt-agnostic target mask, \Method achieves robust attack performance across different prompts.
Our evaluation over multiple datasets and model architectures shows that \Method substantially outperforms existing backdoor attacks. 
For instance, on DAVIS with a 5\% poisoning rate, \Method with Blended achieves 95.3\% ASR under point prompts, while maintaining competitive clean utility with 0.596 mIoU and 0.411 J\&F, and similar trends hold on LVOS.
In addition, we conduct various ablation studies to systematically evaluate the effectiveness and clean utility of \Method, including the effect of the loss function, training stage, and trigger settings, attack performance across different VSFMs, different attack targets, and a study of real-world trigger evaluation. 
Through these experiments, we further demonstrate that \Method consistently achieves effective attack performance across various model architectures and scenarios, revealing potential security vulnerabilities in current mainstream video segmentation foundation models.
Furthermore, we conduct an interpretability analysis to better understand how \Method affects the model's internal representations.
To mitigate the attack, we consider five existing widely-used backdoor defense strategies: Fine-tuning, Pruning~\cite{DBLP:journals/corr/abs-1710-00942}, Spectral Signatures~\cite{DBLP:conf/nips/Tran0M18}, PGBD~\cite{Amula_2025_ICCV}, and STRIP~\cite{DBLP:conf/acsac/GaoXW0RN19}. 
We observe that they are largely ineffective against \Method, highlighting the need for tailored defenses for VSFMs.

Overall, our contributions are summarized as follows:

\begin{itemize}
    \item We propose the first backdoor attack framework (\Method) targeting video segmentation foundation models, which involves a two-stage training paradigm to effectively inject the backdoor pattern.
    \item Built upon \Method, we achieve substantially higher ASR than baseline backdoor attacks across DAVIS and LVOS, with gains of up to around 90 percentage points under point and box prompts, while largely preserving clean-input segmentation quality.
    \item We conduct comprehensive ablation studies to evaluate the effectiveness, generality, and interpretability of \Method across various models and scenarios. In addition, we evaluate several existing defense strategies and find that they are mostly ineffective against \Method, exposing the vulnerability of current VSFMs to structured backdoor threats.
\end{itemize}

%----------------------------------
\section{Preliminary and Related Work}
%----------------------------------

\mypara{Video Object Segmentation (VOS) and Video Segmentation Foundation Models (VSFMs)}
Image segmentation is a crucial task in the field of computer vision and finds wide application in various real-world scenarios, including medical image analysis~\cite{DBLP:conf/bildmed/IsenseePKZJKWKN19,DBLP:conf/icip/SunXWCL024,DBLP:conf/iros/ChenZGL0W0L24}, remote sensing image interpretation~\cite{DBLP:conf/cvpr/LiuMZ0JSJ24,DBLP:conf/cvpr/LiLCBZM025}, and autonomous driving~\cite{DBLP:journals/corr/abs-2401-10153,DBLP:journals/ral/SunYHHW20,DBLP:conf/ivs/TeichmannWZCU18}. 
However, conventional image segmentation methods only operate on static images and struggle to address the challenges posed by temporal variations of objects in videos. 
To better understand and analyze dynamic scenes, video object segmentation (VOS) emerges as a dedicated task that performs pixel-level segmentation of specific targets throughout an entire video sequence~\cite{DBLP:journals/tist/YaoLXZZ20}.

\begin{figure}
    \centering
    \includegraphics[width=1\linewidth]{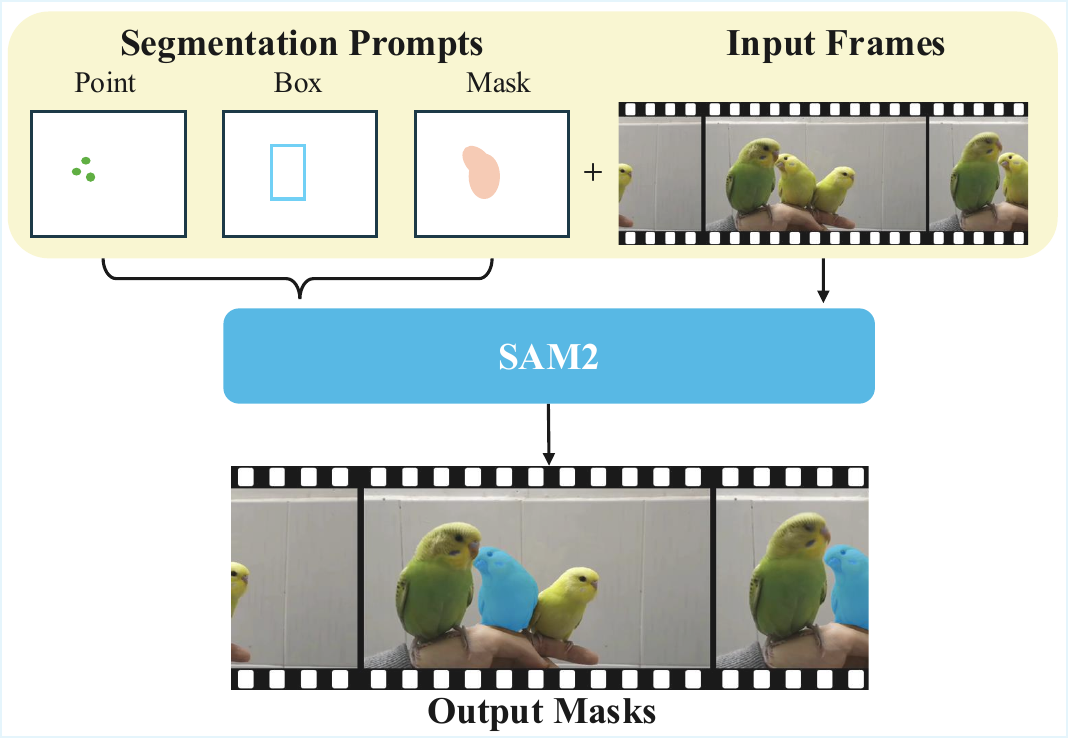}
    \caption{Three types of segmentation prompts for SAM2 (point, box, and mask).}
    \label{fig:prompt_type}
\end{figure}

VOS is typically categorized into two task settings: semi-supervised and unsupervised~\cite{DBLP:journals/tist/YaoLXZZ20}. 
Semi-supervised VOS requires a ground-truth mask of the target object in the first frame as a reference, and the model is expected to continuously track and segment the specified objects in subsequent frames~\cite{DBLP:conf/cvpr/CaellesMPLCG17, DBLP:conf/eccv/JainG14, DBLP:conf/cvpr/PerazziKBSS17}. 
In contrast, unsupervised VOS operates without any prior annotations, and the model must autonomously identify and segment the most salient foreground object in the video~\cite{DBLP:conf/eccv/BroxM10, DBLP:conf/bmvc/FaktorI14,DBLP:conf/iccv/LimHHH13}.
SAM2~\cite{DBLP:conf/iclr/RaviGHHR0KRRGMP25} is one of the most representative and influential foundation models in the field of semi-supervised video object segmentation. 
Built upon the foundation of the SAM~\cite{DBLP:conf/iccv/KirillovMRMRGXW23}, SAM2 extends promptable segmentation to the video domain by introducing a streaming Transformer architecture with a memory bank. 
This design enables SAM2 to store and retrieve object-specific features across frames, allowing for spatio-temporal mask propagation and interactive refinement. 
Unlike static image models, SAM2 processes video frames sequentially while attending to prior object representations, making it well-suited for real-time video segmentation.
Moreover, SAM2 achieves higher accuracy and efficiency in cross-frame video segmentation tasks and consistently outperforms traditional methods~\cite{DBLP:conf/iclr/RaviGHHR0KRRGMP25}.
As shown in \Cref{fig:prompt_type}, similar to SAM, SAM2 also supports multiple types of segmentation prompts to assist in generating mask outputs.
SAM2.1~\cite{sam2} is an enhanced version of SAM2 that offers improved segmentation performance and efficiency.
In addition, SAM2-based models have been widely adopted in various specialized domains. 
For instance, BioSAM2~\cite{DBLP:journals/corr/abs-2408-03286} builds on SAM2 by freezing the prompt encoder and fine-tuning the image encoder and mask decoder on biomedical datasets, thereby improving its capability in medical video analysis. 
Similarly, MedSAM2~\cite{DBLP:journals/corr/abs-2408-00874} is also designed for medical image and video segmentation.
It introduces a self-sorting memory bank mechanism that dynamically selects information embeddings based on confidence and dissimilarity, independent of temporal order. 
This mechanism not only significantly enhances 3D medical image segmentation but also enables one-prompt segmentation for 2D medical images.
Moreover, EdgeTAM, as a mobile-friendly variant of SAM2, targets the main latency bottleneck in SAM2, its memory attention, and replaces it with a lightweight 2D spatial perceiver that encodes dense frame-level memories using a fixed set of learnable queries. 
To preserve spatial locality for dense prediction, the queries are split into global-level and patch-level groups.
Coupled with a distillation pipeline that improves accuracy without adding inference overhead, EdgeTAM achieves substantial efficiency gains while remaining promptable for both segmentation and tracking. 
Concretely, it reports 16 FPS on iPhone 15 Pro Max and up to 22 times speedup over SAM2, with competitive accuracy compared to SAM2. 
These properties make EdgeTAM a practical VSFM for on-device applications at the edge.
However, as these VSFMs see increasing deployment in both industrial and research contexts, it becomes crucial to address the potential security challenges they may encounter.

\mypara{Backdoor Attacks in Video Models}
Backdoor attacks remain a critical topic in the field of AI security~\cite{He2025AISecuritySurvey}, attracting significant attention in both computer vision~\cite{DBLP:journals/corr/abs-1708-06733,DBLP:conf/iccv/LiLWLHL21,DBLP:journals/corr/abs-2111-10991,DBLP:journals/corr/abs-2506-07214,DBLP:journals/corr/abs-2508-15778,guo2026sixdattack} and natural language processing~\cite{DBLP:conf/acl/KuritaMN20,wang2025purity,DBLP:conf/sp/0001CLL0CH025}. 
In a backdoor attack, adversaries manipulate training data by embedding specific triggers or directly releasing models containing hidden triggers. 
When a user input satisfies the trigger condition, the model produces incorrect or harmful outputs~\cite{He2025AISecuritySurvey}.
Backdoor attacks in video-related tasks also receive growing attention. 
Existing research categorizes these attacks based on the targeted task, including video classification~\cite{DBLP:conf/cvpr/ZhaoMZ0CJ20}, video object detection~\cite{DBLP:conf/ijcai/ZhangHWZZWZ024,DBLP:journals/corr/abs-2411-14243}, and video generation~\cite{DBLP:journals/corr/abs-2504-16907}.
For example, Zhao et al.~\cite{DBLP:conf/cvpr/ZhaoMZ0CJ20} propose the use of universal adversarial triggers as backdoor mechanisms in video classification, launching clean-label attacks without modifying ground-truth annotations.
Zhang et al.~\cite{DBLP:conf/ijcai/ZhangHWZZWZ024} identify vulnerabilities in object detection models used in applications such as autonomous driving and surveillance, and introduce a backdoor attack paradigm that renders detectors completely ineffective.
This paradigm includes two attack modes: SPONGE triggers cause the detector to produce an overwhelming number of false positives, while BLINDING triggers prevent the model from detecting actual objects, effectively rendering it blind.
Wang et al.~\cite{DBLP:journals/corr/abs-2504-16907} introduce BadVideo, the first backdoor framework targeting text-to-video diffusion models.
This method leverages the model's tendency to generate redundant content, such as background elements or secondary objects not explicitly specified in the input text.
These elements serve as natural carriers for malicious information, allowing it to be seamlessly embedded into the generated video with high stealth.
Feng et al.~\cite{DBLP:conf/cvpr/FengMZZXT22} propose FIBA, a Frequency-Injection based Backdoor Attack for medical image analysis. 
Instead of overlaying visible spatial patches, FIBA injects the trigger into the amplitude spectrum in the frequency domain by linearly combining the low-frequency components of a trigger image with those of a benign image, while keeping the phase spectrum unchanged. 
This design preserves the spatial semantics of poisoned pixels, enabling stealthy backdoor attacks on both image classification and dense prediction models and helping FIBA bypass many existing defenses.

Although the aforementioned approaches operate in video-input scenarios, the tasks and model architectures they target differ significantly from the VSFMs studied in our work, making them unsuitable for direct transfer.
Therefore, in this study, we adopt several representative image-based backdoor attack methods in computer vision (including BadNet~\cite{DBLP:journals/corr/abs-1708-06733}, WaNet~\cite{DBLP:conf/iclr/NguyenT21}, Blended~\cite{DBLP:conf/iccv/LiLWLHL21}, and FIBA~\cite{DBLP:conf/cvpr/FengMZZXT22}) and apply them to video frames as baselines for comparison. 
These methods are widely adopted and represent diverse attack paradigms, allowing us to assess their effectiveness in video segmentation tasks.
Recent representation-level backdoor attacks manipulate pretrained encoder spaces to implant trigger behaviors~\cite{DBLP:conf/sp/JiaLG22, DBLP:journals/corr/abs-2505-16640}. 
In contrast, \Method studies prompt-driven dense video segmentation, where the attacker must account for prompt-image fusion and mask decoding rather than only image-level representation learning; this requires learning trigger-conditioned mask generation while preserving prompt-driven clean segmentation.

\begin{figure*}[h!]
    \centering
    \includegraphics[width=1.\textwidth]{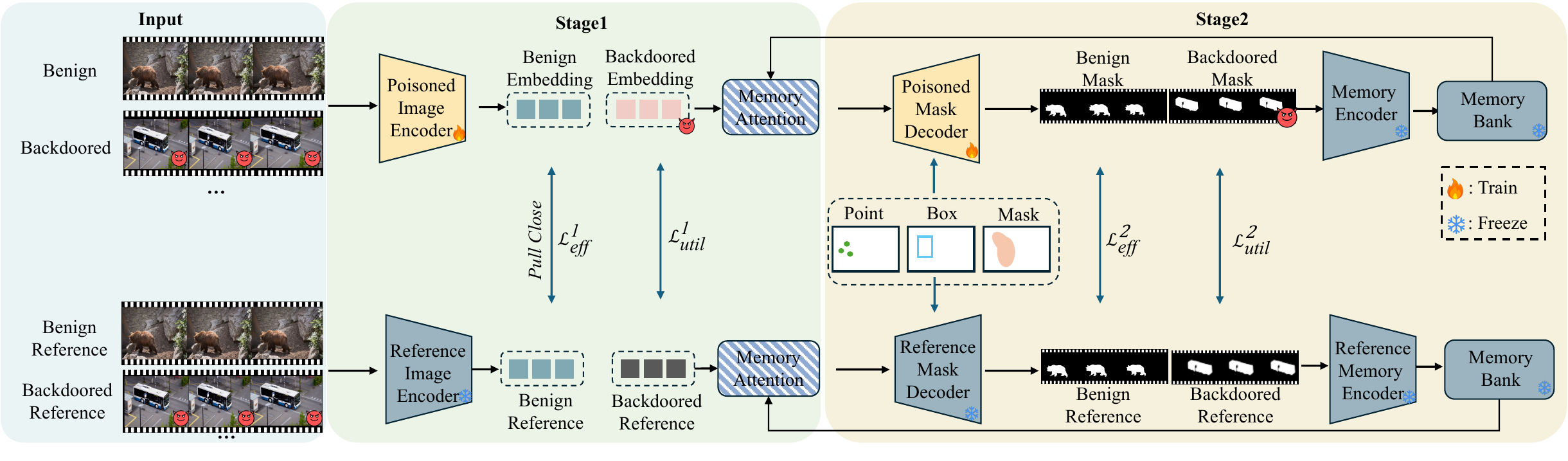}
    \caption{Overview of \Method, a two-stage backdoor attack for VSFMs. Stage 1 separates triggered and clean frame representations by aligning triggered embeddings to a target embedding while preserving clean embeddings. Stage 2 trains the mask decoder to map triggered inputs to the target mask across prompts while maintaining clean segmentation behavior.}
    \label{fig:framework}
\end{figure*}

%----------------------------------
\section{Threat Model}
%----------------------------------

\mypara{Adversary's Goal}
The main goal of the adversary is to publish a backdoored VSFM that outputs a predefined, malicious prediction result when encountering a specific frame containing the trigger pattern while maintaining good performance on the original task with clean videos.
This setting is realistic because, in many domains, the training videos are privacy- or IP-sensitive. 
Providers typically do not release datasets and only publish model weights on platforms such as Hugging Face, making a backdoored release plausible while remaining hard to audit.

\mypara{Adversary's Capabilities}
In this paper, we consider two types of capabilities, i.e., dataset and model.
Regarding the dataset, we assume the adversary can obtain a video segmentation dataset that contains video frames as well as their ground truth segmentation results.
The adversary can further adjust the dataset by introducing backdoor patterns, i.e., triggered frames and modified segmentation to it.
This assumption is realistic as there are numerous video segmentation datasets from companies~\cite{DBLP:journals/corr/abs-1809-03327,DBLP:conf/eccv/MeiZYYQCK22}, open-source platforms~\cite{DBLP:conf/iccv/HongCLZGCZ23,DBLP:conf/iccv/LiKHTR13}, and competitions~\cite{DBLP:journals/corr/Pont-TusetPCASG17,cvpr2018_autonomous_driving}.
Regarding the model, we assume the adversary can obtain a pre-trained VSFM and fine-tune it with the previous dataset and custom loss functions.
This assumption also holds since there are various publicly accessible pre-trained VSFMs, and the fine-tuning costs are much smaller than the pre-training procedure, which is affordable to the adversary.

% ----------------------------------------------------
\section{Methodology}
% ----------------------------------------------------
\label{sec:method}

Modern prompt-driven video segmentation models, such as SAM2, generate masks by combining frame embeddings from the image encoder with prompt embeddings in the decoder.
In this work, unless otherwise specified, we consider an object-disappearance backdoor objective. 
When the trigger is absent, the model should preserve normal video segmentation behavior, whereas when the trigger appears, the model should output an all-zero target mask for the prompted object across different prompt types.
We also study a non-zero deformation target in \Cref{sec:ablation_study}.
In our experiments, directly applying traditional single-stage backdoor training, such as BadNet, to SAM2 leads to close to 0\% ASR.
We further conduct a gradient-conflict analysis to analyze why the traditional backdoor methods failed in \Cref{sec:attack_performance}.
Together with the strong spatial constraints provided by point, box, or mask prompts, the decoder is encouraged to fit clean and triggered inputs similarly instead of learning a stable trigger-conditioned target-mask behavior.

To overcome this, we present \Method, a two-stage backdoor attack against prompt-driven video segmentation models.
The detailed pipeline of \Method is shown in \Cref{fig:framework}.
Specifically, stage 1 fine-tunes the image encoder to map the triggered frame embedding into a predefined target embedding while keeping the clean frame embedding unchanged (aligned with a reference clean model).
Stage 2 updates the decoder to render the triggered frame's embedding into the target mask across different prompts while keeping the clean frame mask unchanged as well. 
Below, we discuss the detailed design in both stages.

\mypara{Stage 1}
In stage 1, we only update the image encoder \(f\) and freeze all other modules.
We define \(f'\) to be the image encoder of a reference model trained on clean data, 
\(\mathcal{D}_{\mathrm{trig}}\) and \(\mathcal{D}_{\mathrm{clean}}\) to be the sets of triggered and clean training frames, respectively, 
and \(x_{target}\) to be an attacker predefined target image with target embedding \(f'(x_{target})\).
To meet the effectiveness and utility goals, we introduce two losses.
The effectiveness loss pulls embeddings of triggered frames toward the predefined target embedding \(f'(x_{target})\):
\begin{equation}
\mathcal{L}^{1}_{\mathrm{eff}}
=\frac{1}{|\mathcal{D}_{\mathrm{trig}}|}
\sum_{x \in \mathcal{D}_{\mathrm{trig}}}
\big\| f(x) - f'(x_{target}) \big\|_2^2 .
\end{equation}

The utility loss preserves clean performance by aligning clean embeddings with those from the reference encoder:
\begin{equation}
\mathcal{L}^{1}_{\mathrm{util}}
=\frac{1}{|\mathcal{D}_{\mathrm{clean}}|}
\sum_{x \in \mathcal{D}_{\mathrm{clean}}}
\big\| f(x) - f'(x) \big\|_2^2 .
\end{equation}

The final loss in stage 1 can be defined as a weighted sum of the two losses mentioned above:
\begin{equation}
\mathcal{L}_{stage1}
=\lambda_{\mathrm{1}}\mathcal{L}^{1}_{\mathrm{eff}}
+\mathcal{L}^{1}_{\mathrm{util}},
\end{equation}
where \(\lambda_{\mathrm{1}}\) is a weighting coefficient.
During training, we update only \(f\) to achieve the above objectives: the effectiveness loss unifies triggered-frame features at a target embedding, while the utility loss preserves performance on clean data.

\mypara{Stage 2}
This stage only updates the mask decoder \(g\) and freezes all other modules.
Let \(g'\) denote the mask decoder of the clean reference model, and let \(\mathcal{P}\) be the set of supported prompts, including point, box, and mask prompts in our experiments.
The attacker-predefined target mask is \(m_{\text{target}}\in\mathcal{C}^{H\times W}\), where each pixel stores a class index from the set \(\mathcal{C}\), and \(H,W\) denote the mask height and width.
Given the image encoder \(f\) and a prompt encoder \(h\), we define the per-pixel foreground probability by a sigmoid activation: \(p(x,r)=\sigma\!\big(g(f(x),\,h(r))\big)\) for a frame \(x\) and a prompt \(r\in\mathcal{P}\). 
To meet these two goals, we use an effectiveness loss on triggered frames and a utility loss on clean frames.
The effectiveness loss supervises all triggered-frame predictions toward the shared target across prompts using \(\mathcal{L}_{\mathrm{BCE}}\) and \(\mathcal{L}_{\mathrm{DICE}}\):
\begin{equation}
\label{eq:s2-eff}
\begin{aligned}
\mathcal{L}^{2}_{\mathrm{eff}}
&=\frac{1}{|\mathcal{D}_{\mathrm{trig}}|\,|\mathcal{P}|}
  \sum_{x\in\mathcal{D}_{\mathrm{trig}}}
  \sum_{r\in\mathcal{P}}
  \Big[
    \mathcal{L}_{\mathrm{BCE}}\big(\sigma\!\big(g(f(x),\,h(r))\big)\,\,q\big) \\
&\qquad\qquad
  + \mathcal{L}_{\mathrm{DICE}}\big(\sigma\!\big(g(f(x),\,h(r))\big)\,\,q\big)
  \Big],
\end{aligned}
\end{equation}
where \(q=\Pi_{\text{union}}(m_{\text{target}})\in\{0,1\}^{H\times W}\) takes value \(1\) if a pixel belongs to any foreground class and \(0\) otherwise.
Let \(n=H\cdot W\) be the number of pixels; the BCE and Dice losses are defined as:
\begin{equation}
\label{eq:s2-bce}
\mathcal{L}_{\mathrm{BCE}}(p,q)
=-\frac{1}{n}\sum_{i=1}^{n}\Big[q_i\log p_i+(1-q_i)\log(1-p_i)\Big],
\end{equation}
\begin{equation}
\label{eq:s2-dice}
\mathcal{L}_{\mathrm{DICE}}(p,q)
=1-\frac{2\sum_{i=1}^{n}p_i q_i+\varepsilon}{\sum_{i=1}^{n}p_i+\sum_{i=1}^{n}q_i+\varepsilon},
\end{equation}
where a small constant \(\varepsilon>0\) is set for numerical stability.

The utility loss preserves performance on clean data by aligning decoder logits with the reference decoder on clean frames:
\begin{equation}
\label{eq:s2-util}
\begin{aligned}
\mathcal{L}^{2}_{\mathrm{util}}
&=\frac{1}{|\mathcal{D}_{\mathrm{clean}}|\,|\mathcal{P}|}
  \sum_{x\in\mathcal{D}_{\mathrm{clean}}}
  \sum_{r\in\mathcal{P}} \\
&\qquad \times\big\|\,g\big(f(x),h(r)\big)-g'\big(f(x),h(r)\big)\,\big\|_2^2 .
\end{aligned}
\end{equation}

The final stage 2 objective is the weighted sum:
\begin{equation}
\label{eq:s2-obj}
\mathcal{L}_{\text{stage2}}
=\mathcal{L}^{2}_{\mathrm{eff}}
+\lambda_{\mathrm{2}}\,\mathcal{L}^{2}_{\mathrm{util}},
\end{equation}
where \(\lambda_{\mathrm{2}}\)\ is a weighting coefficient.
During stage 2 training, we optimize only the mask decoder \(g\) to minimize \(\mathcal{L}_{\text{stage2}}\).
The effectiveness loss drives triggered predictions toward the shared target across prompts, while the utility loss preserves clean accuracy by aligning the current decoder with the reference decoder \(g'\).

%----------------------------------
\section{Experimental Settings}
%----------------------------------
\label{sec:setting}

\mypara{Datasets and Models}
We adopt the following two public datasets to conduct experimental evaluation: LVOS~\cite{DBLP:conf/iccv/HongCLZGCZ23} and DAVIS-2017 (DAVIS)~\cite{DBLP:journals/corr/Pont-TusetPCASG17}.
LVOS targets long-term video object segmentation and contains 720 videos with 296,401 frames and 407,945 pixel-level annotations. 
We adopt the official split of 420 training, 140 validation, and 160 test videos.
DAVIS is a densely annotated video segmentation challenge dataset, which contains 150 video sequences, a total of 10,459 frames of annotations, and 376 objects, and we follow the default training and test set divisions provided by the dataset.
For the test sets of both datasets, we prepare two versions: a fully triggered split, in which every frame contains the trigger to evaluate backdoor performance, and a clean split with no triggers to evaluate clean performance.
For victim models, we use the pre-trained SAM2-base-plus backbone~\cite{DBLP:conf/iclr/RaviGHHR0KRRGMP25} for the comparison with the baseline backdoor attacks, and we further apply \Method to other VSFMs to evaluate its effectiveness, including MedSAM2~\cite{DBLP:journals/corr/abs-2408-00874}, SAM2-Long~\cite{DBLP:journals/corr/abs-2410-16268}, BioSAM2~\cite{DBLP:journals/corr/abs-2408-03286}, and EdgeTAM~\cite{DBLP:conf/cvpr/ZhouZXSXWK0LCS25}.

\mypara{Baseline Backdoor Attacks Deployment}
In our experiments, we consider four baseline backdoor attacks: BadNet~\cite{DBLP:journals/corr/abs-1708-06733}, Blended~\cite{DBLP:conf/iccv/LiLWLHL21}, WaNet~\cite{DBLP:conf/iclr/NguyenT21}, and FIBA~\cite{DBLP:conf/cvpr/FengMZZXT22}.
These methods cover both visible triggers, such as BadNet and Blended, and invisible triggers, including WaNet and FIBA.
Models take video frames as input and produce segmentation masks as output.
Below, we detail the mask targets and the trigger injection procedures.
The input poisoned video frames are paired with an attack target mask. 
In the main experiments, an all-zero mask is used as the output mask. 
In other ablation experiments, we also use a different attack target mask, specifically a central circle mask.
Clean frames retain their original ground truth masks.
Regarding trigger injection settings, BadNet places a \(40\times40\) solid-color patch at the bottom-right corner with an \(8\)-pixel margin.
This keeps the footprint in the same ballpark as common BadNet settings~\cite{DBLP:journals/corr/abs-1708-06733}.
For Blended, we follow the popular setting~\cite{DBLP:conf/iccv/LiLWLHL21} that blends a small random texture into a random-texture square at the bottom right with side length equal to 18\% of the image short side, with \(\alpha=0.18\). 
WaNet uses the default setting~\cite{DBLP:conf/iclr/NguyenT21} that applies a fixed smooth displacement field generated with a Gaussian kernel of \(101\) and a maximum displacement of \(0.01\) times the short side.
FIBA injects low-frequency amplitude in the Fourier domain by mixing within a central window of fraction \(0.06\) using a blending coefficient of \(0.25\), and this setting is followed by~\cite{DBLP:conf/cvpr/FengMZZXT22}.

\mypara{Implementation Details}
Regarding backdoored model training, we train \Method for \(2\) epochs in stage 1 and \(3\) epochs in stage 2; all other methods, including the reference model, are trained for \(5\) epochs.
For both \Method and all baseline attacks, we follow the default SAM2 training setting~\cite{sam2}, using AdamW with a batch size of \(1\) and a learning rate of \(1.0\times10^{-5}\).
For \Method, we set the training hyperparameters as \(\lambda_{\mathrm{1}}=\lambda_{\mathrm{2}}=1\) in both stage 1 and stage 2.
All experiments are conducted on a server equipped with eight NVIDIA L20 GPUs.

\mypara{Training Cost}
We measure the computational overhead on DAVIS. Vanilla SAM2 fine-tuning takes 98 minutes, while \Method takes 167 minutes under the same setting, resulting in a 1.7$\times$ overhead.
The additional cost mainly comes from the frozen reference-model forward pass used for clean-behavior alignment.
Since the dominant cost remains standard VSFM fine-tuning, the overhead is a constant-factor increase under our threat model.

\mypara{Evaluation Metric}
To evaluate the utility of segmentation performance, we adopt two common metrics: mean Intersection over Union (mIoU) and the J\&F measure~\cite{DBLP:journals/corr/Pont-TusetPCASG17}.
Given a video of $N$ frames, let $P_i$ and $G_i$ denote the predicted and ground-truth masks for the objects in the $i$-th frame. 
The mIoU is defined as:

\begin{equation}
\text{mIoU} = \frac{1}{N} \sum_{i=1}^{N} \frac{|P_i \cap G_i|}{|P_i \cup G_i|},
\label{equ:miou}
\end{equation}
which reflects the average overlap between predicted and ground-truth masks across frames.
Following previous work for a more comprehensive evaluation~\cite{DBLP:journals/corr/abs-2503-12781}, we also use the J\&F metric from the DAVIS benchmark, which combines region similarity (Jaccard index) and contour accuracy (F1 Score):

\begin{equation}
\text{J\&F}=\frac12\Bigl(
  \underbrace{mIoU}_{\substack{\textbf{J: Region Similarity}\\(Jaccard\ Index)}}
  \;\;+\;\; 
  \underbrace{\frac1N\sum_{i=1}^{N}\frac{2\,\text{Precision}_i\,\text{Recall}_i}
                      {\text{Precision}_i+\text{Recall}_i}}
              _{\substack{\textbf{F: Contour Accuracy}\\(F1\ Score)}}
\Bigr),
\end{equation}
where $\text{Precision}_i = \frac{|P_i \cap G_i|}{|P_i|}$ and $\text{Recall}_i = \frac{|P_i \cap G_i|}{|G_i|}$.

To quantify the effectiveness of the backdoor attack, we use the attack success rate (ASR) as the evaluation metric.
We consider two cases for the measurement.
If the attack target mask is all-zero, we count the ratio of video frames whose predictions are also all-zero:
\begin{equation}
\mathrm{ASR} \;=\; \frac{\big|\{\, i \in \{1,\ldots,N\} : \lVert P_i\rVert_1 = 0 \,\}\big|}{N}.
\end{equation}
If the attack target is non-zero, we average the overlap between the prediction and the target mask over video frames:
\begin{equation}
\mathrm{ASR} \;=\; \frac{1}{N} \sum_{i=1}^{N} \mathrm{IoU}\!\left(P_i, T_i\right),
\end{equation}
where \(T_i\) denotes the attack target mask for the \(i\)-th frame.
Overall, ASR measures target-specific attack success: for an all-zero target, it is the share of frames predicted as all zero; for a non-zero target, the sequence-averaged IoU between predictions and the target masks.

%----------------------------------
\section{Evaluation}
%----------------------------------

\begin{table*}[t]
\centering
\small
\setlength{\tabcolsep}{3.2pt}
\renewcommand{\arraystretch}{1.12}
\caption{Performance of backdoor attacks on SAM2 with a 5\% poisoning rate across datasets and prompts.}
\label{tab:backdoor_sam2_main}
\begin{tabular*}{\textwidth}{@{\extracolsep{\fill}}llccc ccc ccc@{}}
\toprule
\multicolumn{1}{c}{\multirow{2}{*}{\textbf{Dataset}}} &
\multicolumn{1}{c}{\multirow{2}{*}{\textbf{Method}}} &
\multicolumn{3}{c}{\textbf{Point}} &
\multicolumn{3}{c}{\textbf{Box}} &
\multicolumn{3}{c}{\textbf{Mask}} \\
\cmidrule(lr){3-5}\cmidrule(lr){6-8}\cmidrule(lr){9-11}
& & \textbf{mIoU} & \textbf{J\&F} & \textbf{ASR~(\%)} 
  & \textbf{mIoU} & \textbf{J\&F} & \textbf{ASR~(\%)} 
  & \textbf{mIoU} & \textbf{J\&F} & \textbf{ASR~(\%)} \\
\midrule
\multirow{9}{*}{DAVIS}
& Clean          & 0.642 & 0.526 & 2.1 & 0.902 & 0.729 & 0.0 & 0.909 & 0.740 & 0.0 \\
\cmidrule(lr){2-11}
& BadNet         & 0.548 & 0.368 & 3.5 & 0.845 & 0.579 & 0.0 & 0.891 & 0.638 & 0.0 \\
& \Method with BadNet
& 0.556 & 0.377 & \textbf{93.0}
& 0.852 & 0.586 & \textbf{92.2}
& 0.895 & 0.643 & \textbf{48.3} \\
\cmidrule(lr){2-11}
& Blended        & 0.589 & 0.404 & 3.4 & 0.839 & 0.574 & 0.0 & 0.891 & 0.640 & 0.0 \\
& \Method with Blended
& 0.596 & 0.411 & \textbf{95.3}
& 0.846 & 0.583 & \textbf{94.1}
& 0.895 & 0.643 & \textbf{66.1} \\
\cmidrule(lr){2-11}
& FIBA           & 0.582 & 0.399 & 4.0 & 0.853 & 0.589 & 0.0 & 0.896 & 0.644 & 0.0 \\
& \Method with FIBA
& 0.578 & 0.395 & \textbf{64.9}
& 0.849 & 0.584 & \textbf{52.2}
& 0.893 & 0.640 & \textbf{26.3} \\
\cmidrule(lr){2-11}
& WaNet          & 0.560 & 0.384 & 3.5 & 0.837 & 0.573 & 0.0 & 0.889 & 0.639 & 0.0 \\
& \Method with WaNet
& 0.567 & 0.390 & \textbf{85.5}
& 0.844 & 0.580 & \textbf{87.0}
& 0.892 & 0.642 & \textbf{43.6} \\
\midrule
\multirow{9}{*}{LVOS}
& Clean          & 0.610 & 0.474 & 0.0 & 0.897 & 0.711 & 0.0 & 0.912 & 0.829 & 0.0 \\
\cmidrule(lr){2-11}
& BadNet         & 0.568 & 0.426 & 0.0 & 0.862 & 0.621 & 0.0 & 0.914 & 0.822 & 0.0 \\
& \Method with BadNet
& 0.575 & 0.432 & \textbf{94.2}
& 0.875 & 0.629 & \textbf{92.5}
& 0.920 & 0.827 & \textbf{46.0} \\
\cmidrule(lr){2-11}
& Blended        & 0.584 & 0.426 & 0.0 & 0.835 & 0.631 & 0.0 & 0.887 & 0.824 & 0.0 \\
& \Method with Blended
& 0.591 & 0.433 & \textbf{95.1}
& 0.842 & 0.639 & \textbf{94.3}
& 0.894 & 0.829 & \textbf{64.0} \\
\cmidrule(lr){2-11}
& FIBA           & 0.576 & 0.434 & 0.0 & 0.839 & 0.646 & 0.0 & 0.895 & 0.831 & 0.0 \\
& \Method with FIBA
& 0.583 & 0.441 & \textbf{63.1}
& 0.846 & 0.654 & \textbf{52.6}
& 0.892 & 0.828 & \textbf{25.3} \\
\cmidrule(lr){2-11}
& WaNet          & 0.566 & 0.430 & 0.0 & 0.846 & 0.670 & 0.0 & 0.887 & 0.826 & 0.0 \\
& \Method with WaNet
& 0.573 & 0.437 & \textbf{82.3}
& 0.840 & 0.664 & \textbf{83.1}
& 0.891 & 0.830 & \textbf{42.3} \\
\bottomrule
\end{tabular*}
\end{table*}

%----------------------------------
\subsection{Attack Performance}
%----------------------------------
\label{sec:attack_performance}

\begin{figure*}
    \centering
    \includegraphics[width=1\linewidth]{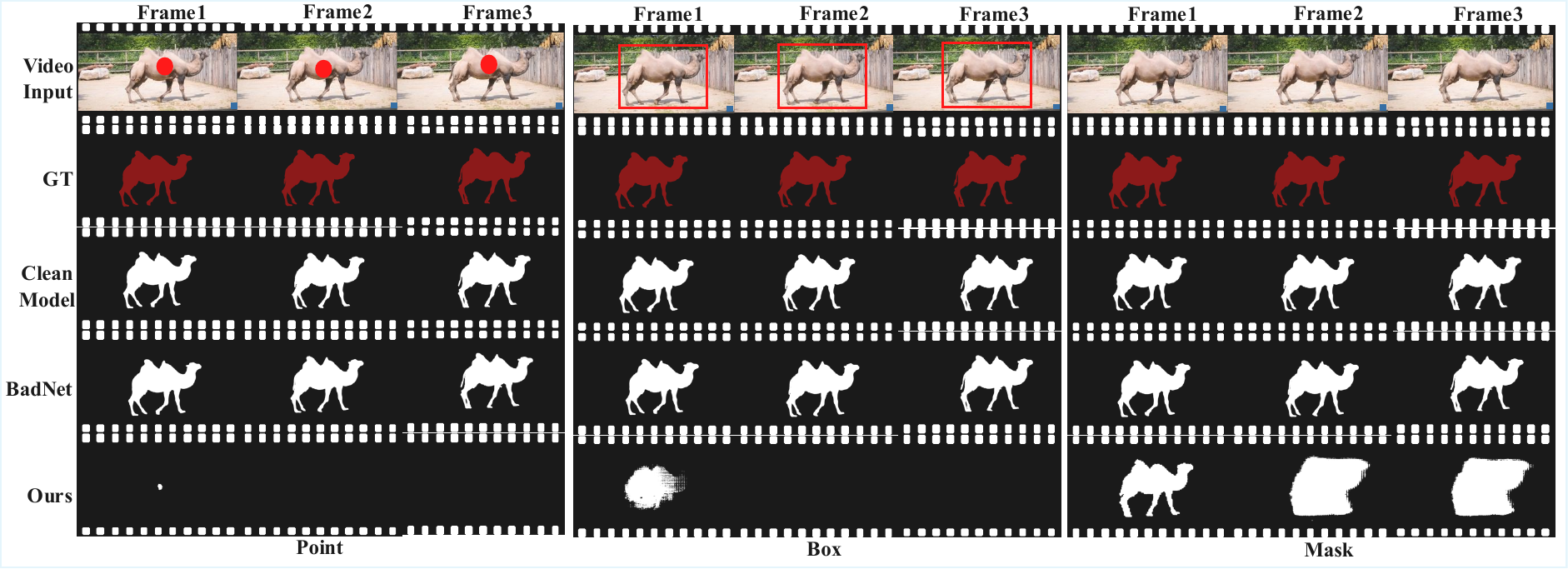}
    \caption{Visualization result under three prompt types (Point, Box, and Mask). Each block shows three consecutive frames (top row: video input with prompt); the rows below give ground truth, SAM2 output on clean input, output under the BadNet baseline, and output under our method.}
    \label{fig:badcase}
\end{figure*}

\mypara{Comparison of Different Baseline Attacks}
To evaluate the attack performance, we compare \Method with four representative visual backdoor attacks on SAM2 (BadNet~\cite{DBLP:journals/corr/abs-1708-06733}, Blended~\cite{DBLP:conf/iccv/LiLWLHL21}, WaNet~\cite{DBLP:conf/iclr/NguyenT21}, and FIBA~\cite{DBLP:conf/cvpr/FengMZZXT22}) on two datasets, DAVIS~\cite{DBLP:journals/corr/Pont-TusetPCASG17} and LVOS~\cite{DBLP:conf/iccv/HongCLZGCZ23}.
Since \Method is not tied to any specific trigger design, we apply \Method as a training framework on top of each baseline, without altering implementation details such as the trigger's shape/pattern, blending ratio, spatial position and size, geometric transformation strategy, color intensity, or occurrence frequency.
All methods are evaluated under three prompt types (point, box, and mask) with ASR, mIoU, and J\&F as the evaluation metrics.
As shown in~\Cref{tab:backdoor_sam2_main}, taking DAVIS with SAM2 at a 5\% poisoning rate as an example, the original baseline attacks achieve limited attack success under all prompt types, while \Method substantially increases ASR when applied to each baseline attack.
Specifically, under the point prompt, Blended only reaches 3.4\% ASR, whereas \Method with Blended improves ASR to 95.3\%.
Meanwhile, the clean utility remains comparable, with mIoU changing from 0.589 to 0.596 and J\&F changing from 0.404 to 0.411.
Similarly, all baseline attacks achieve 0\% ASR under the box prompt, whereas \Method with Blended reaches the highest ASR of 94.1\%.
Under the mask prompt, all baseline attacks still achieve 0\% ASR, while \Method obtains a best ASR of 66.1\%.
The relatively lower ASR under the mask prompt is expected, because the mask prompt supplies the ground-truth mask from the first frame as input to the VSFM, which constrains the model to segment that region and therefore suppresses attack success compared to point and box prompts.
A similar trend appears on the LVOS dataset, as shown in~\Cref{tab:backdoor_sam2_main}.
For example, under the point prompt, \Method with BadNet attains 94.2\% ASR, improving over the BadNet baseline by 94.2 percentage points.
At the same time, its clean utility remains close to the BadNet baseline, with mIoU changing from 0.568 to 0.575 and J\&F changing from 0.426 to 0.432, while remaining within 0.035 and 0.042 of the clean model, respectively.
For the box prompt, \Method with Blended achieves 94.3\% ASR, while the corresponding baseline has 0\% ASR.
For the mask prompt, \Method with Blended achieves the highest ASR of 64.0\%, again demonstrating that \Method can improve attack success even under stronger prompt guidance.
A visualization example of \Method is shown in~\Cref{fig:badcase}.

Overall, across DAVIS and LVOS and for all three prompt types, \Method delivers substantial ASR improvements over the original baseline attacks.
At the same time, the clean utility metrics mIoU and J\&F remain comparable to the corresponding baselines, with only moderate fluctuations across different triggers and prompt types.
These results indicate that \Method enhances attack success while largely preserving the quality of clean-frame segmentation.

\begin{figure}[t]
    \centering
    \begin{subfigure}[t]{0.5\linewidth}
        \centering
        \includegraphics[width=\linewidth]{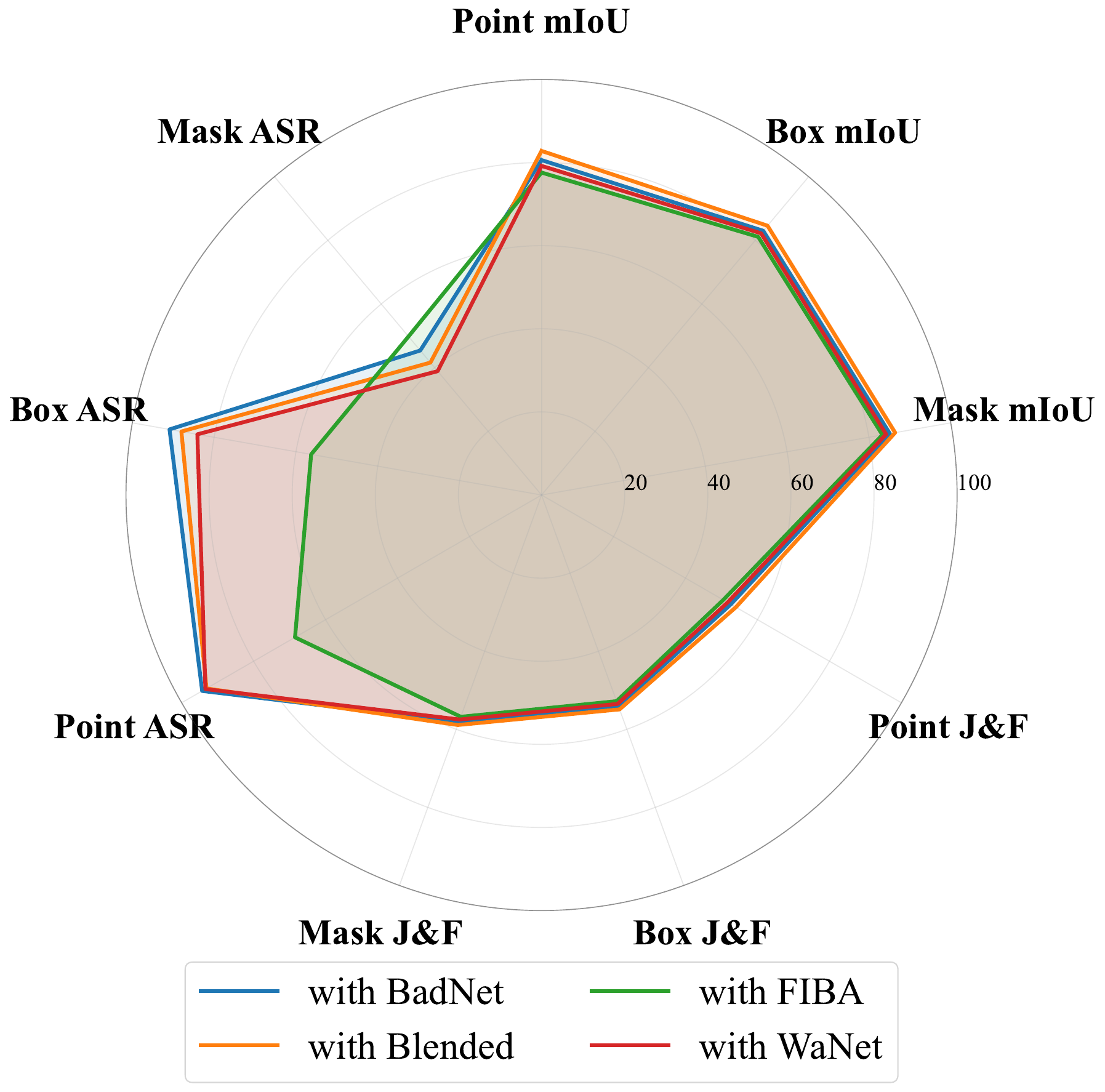}
        \caption{BioSAM2}
    \end{subfigure}\hfill
    \begin{subfigure}[t]{0.5\linewidth}
        \centering
        \includegraphics[width=\linewidth]{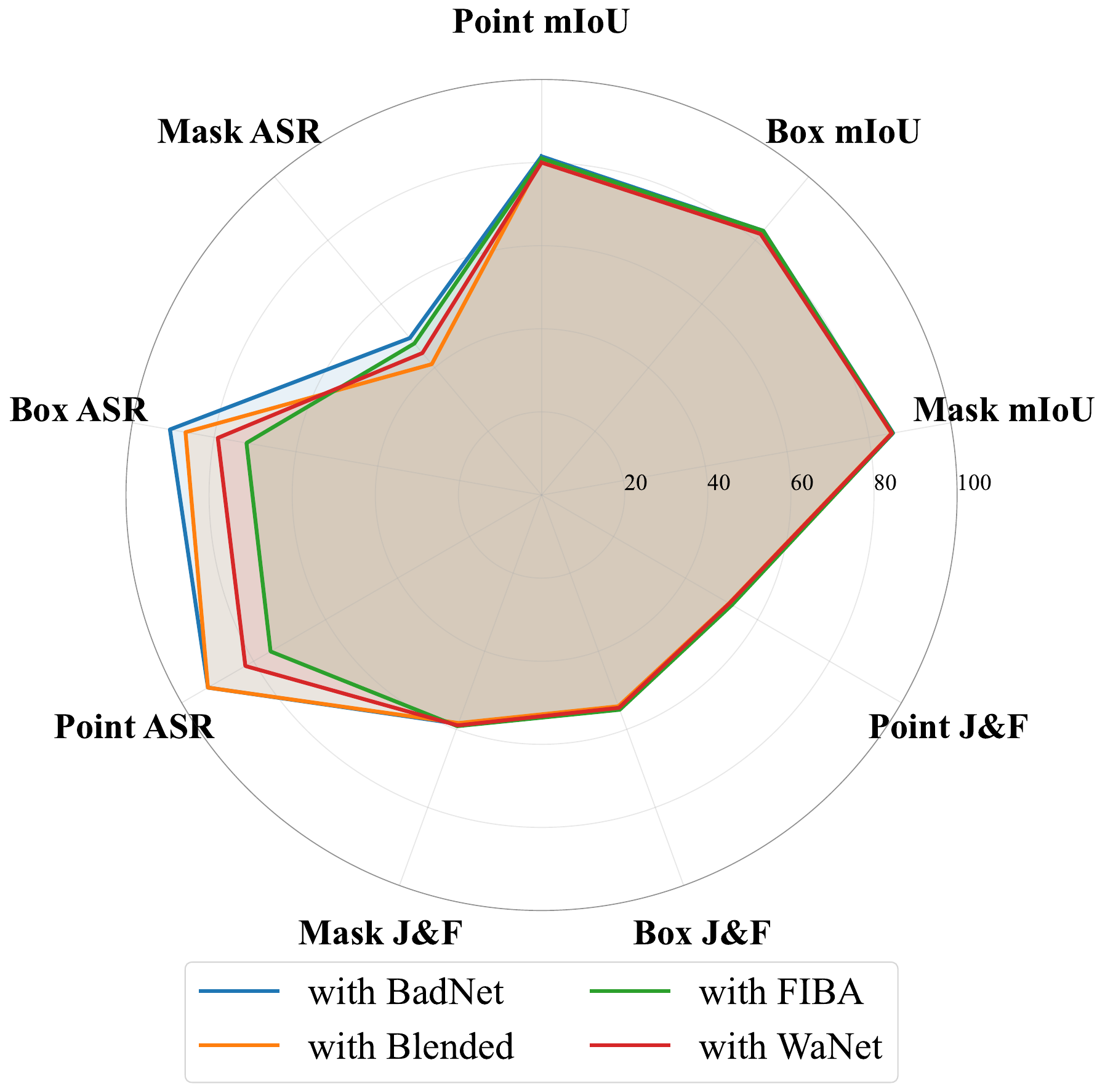}
        \caption{MedSAM2}
    \end{subfigure}

    \vspace{0.6em}

    \begin{subfigure}[t]{0.5\linewidth}
        \centering
        \includegraphics[width=\linewidth]{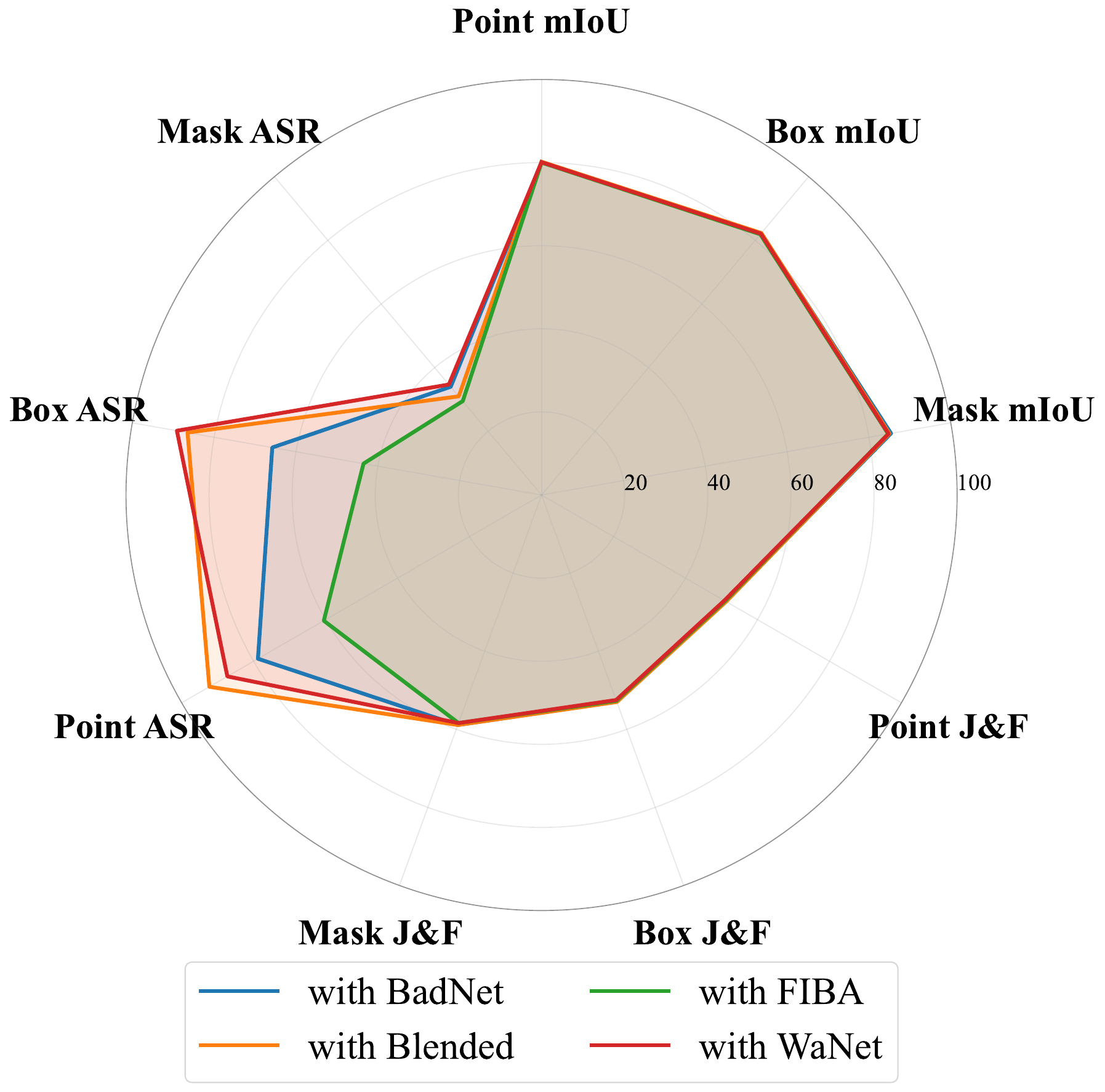}
        \caption{SAM2-Long}
    \end{subfigure}\hfill
    \begin{subfigure}[t]{0.5\linewidth}
        \centering
        \includegraphics[width=\linewidth]{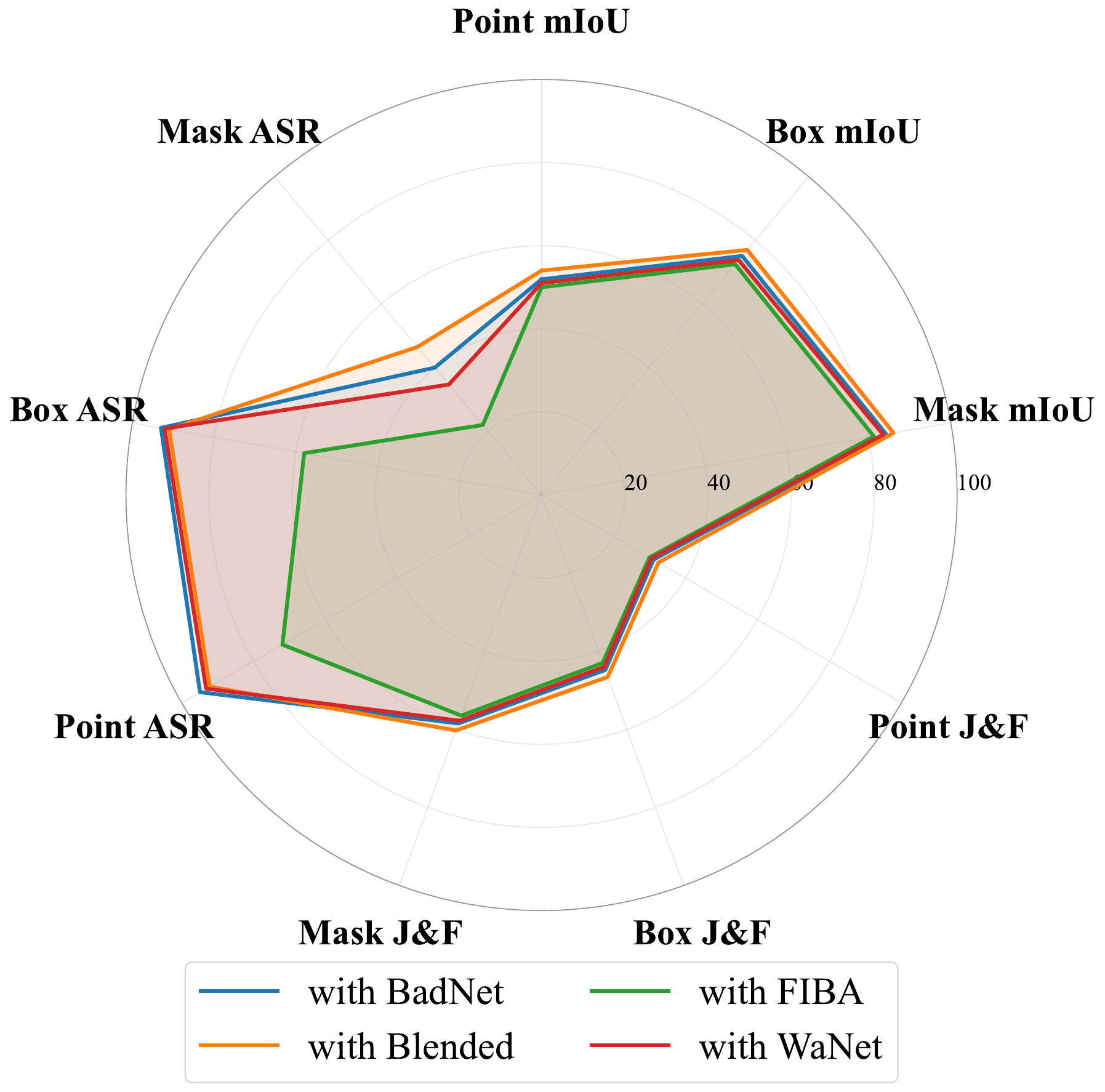}
        \caption{EdgeTAM}
    \end{subfigure}

    \caption{Backdoor performance for BioSAM2, MedSAM2, SAM2-Long, and EdgeTAM on the DAVIS dataset at a 5\% poisoning rate.}
    \label{fig:davis_radar_newvsfms}
\end{figure}

\begin{figure}[t]
    \centering
    \begin{subfigure}[t]{0.5\linewidth}
        \centering
        \includegraphics[width=\linewidth]{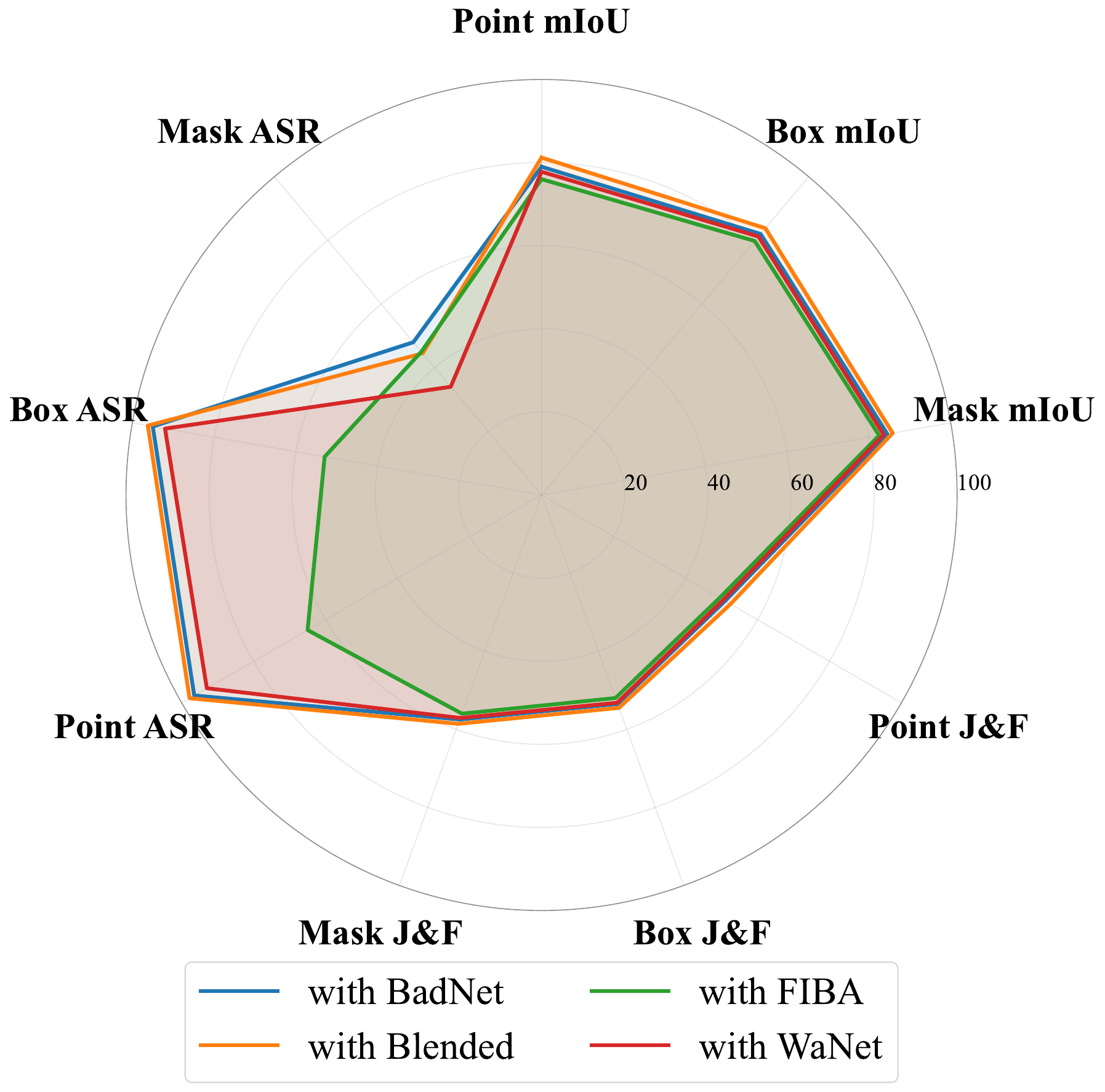}
        \caption{BioSAM2}
    \end{subfigure}\hfill
    \begin{subfigure}[t]{0.5\linewidth}
        \centering
        \includegraphics[width=\linewidth]{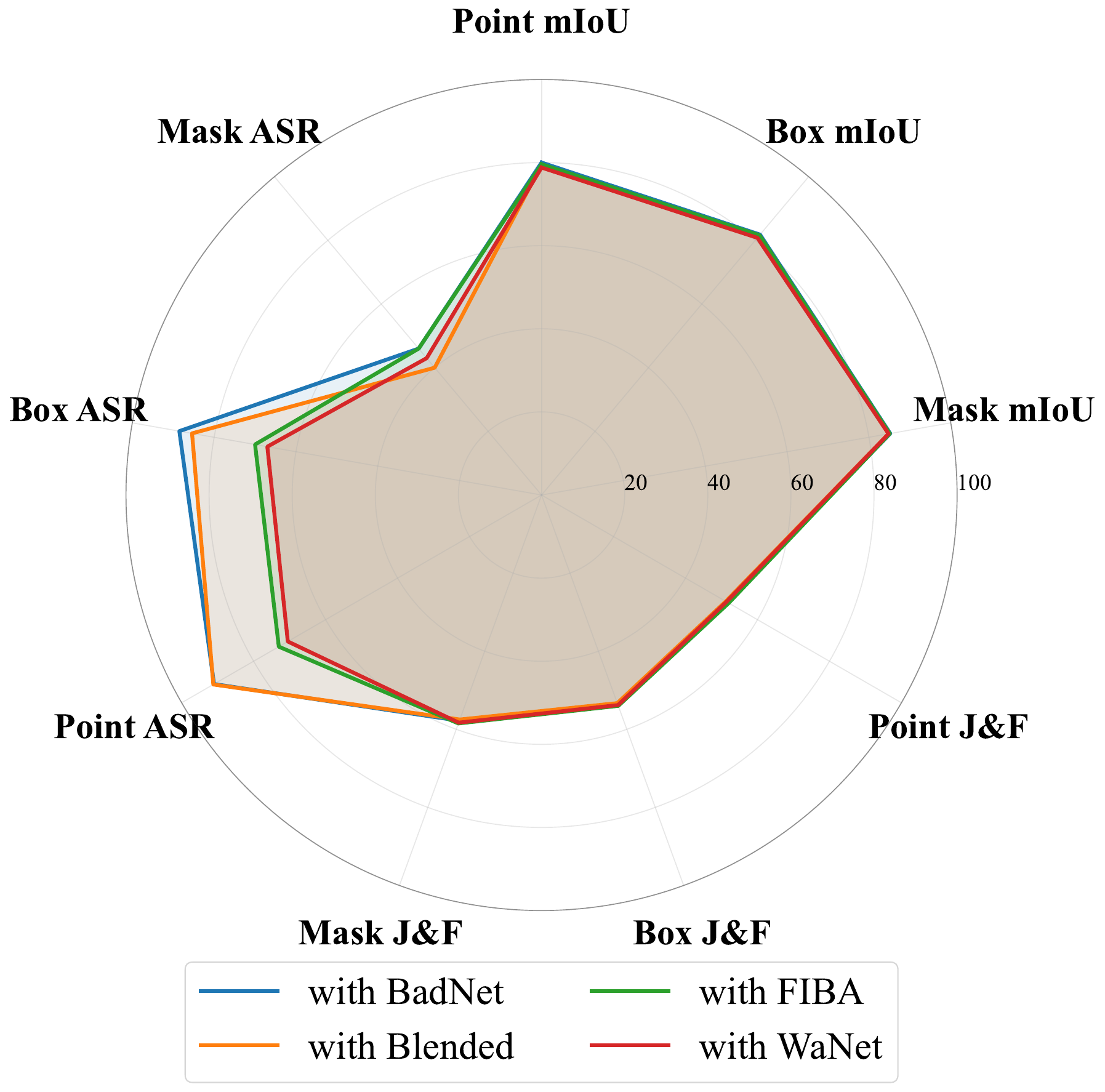}
        \caption{MedSAM2}
    \end{subfigure}

    \vspace{0.6em}

    \begin{subfigure}[t]{0.5\linewidth}
        \centering
        \includegraphics[width=\linewidth]{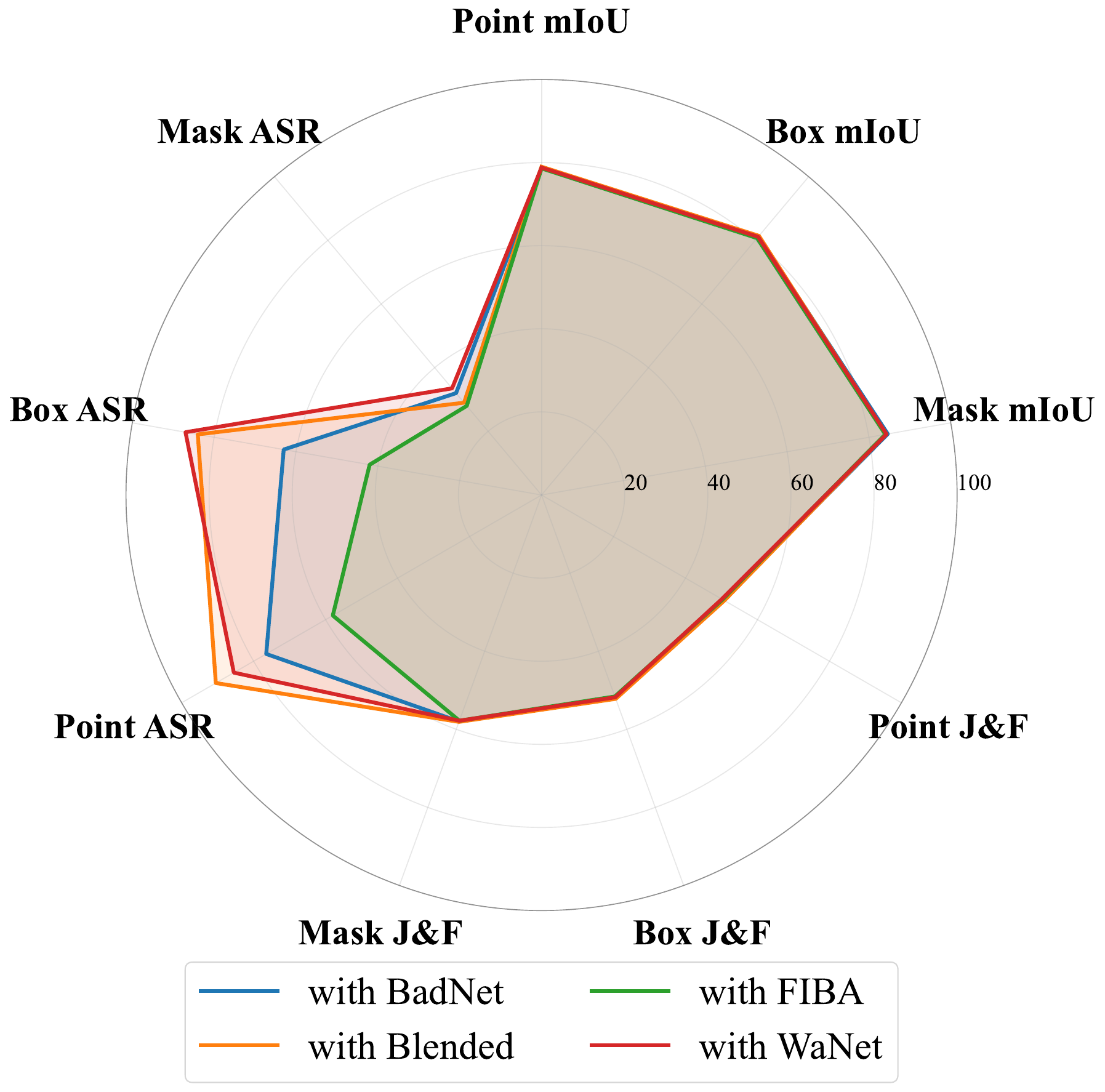}
        \caption{SAM2-Long}
    \end{subfigure}\hfill
    \begin{subfigure}[t]{0.5\linewidth}
        \centering
        \includegraphics[width=\linewidth]{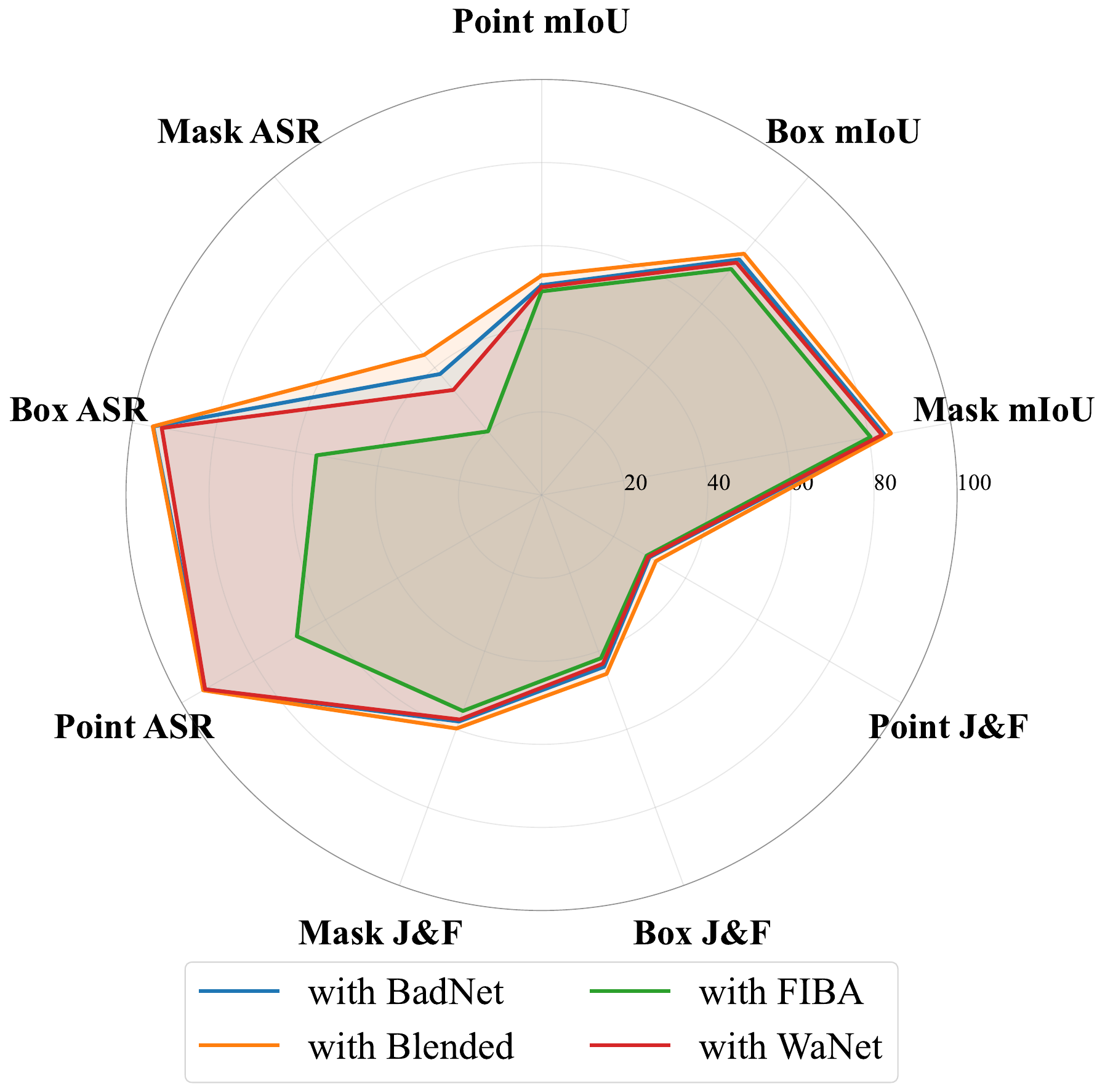}
        \caption{EdgeTAM}
    \end{subfigure}

    \caption{Backdoor performance for BioSAM2, MedSAM2, SAM2-Long, and EdgeTAM on the LVOS dataset at a 5\% poisoning rate.}
    \label{fig:lvos_radar_newvsfms}
\end{figure}

\mypara{Attack Performance Across Different VSFMs}
We now extend our evaluation beyond SAM2 to four additional video segmentation models: MedSAM2~\cite{DBLP:journals/corr/abs-2408-00874}, SAM2-Long~\cite{DBLP:journals/corr/abs-2410-16268}, BioSAM2~\cite{DBLP:journals/corr/abs-2408-03286}, and EdgeTAM~\cite{DBLP:conf/cvpr/ZhouZXSXWK0LCS25}.
As summarized in \Cref{fig:davis_radar_newvsfms,fig:lvos_radar_newvsfms}, \Method shows a similar pattern on these VSFMs.
In EdgeTAM with DAVIS, \Method with BadNet achieves the highest ASR on point and box, reaching 93.2\% and 92.1\% respectively, while keeping utility competitive, for example, mIoU and J\&F are 0.510 and 0.305 on point and 0.738 and 0.440 on box, indicating competitive clean utility.
In BioSAM2 with DAVIS, the mask-prompt results also follow this trend: \Method with Blended achieves the highest ASR of 88.6\%, with mIoU 0.864 and J\&F 0.589, indicating strong backdoor performance while preserving segmentation quality.
On MedSAM2 with DAVIS, \Method with BadNet delivers solid performance across prompts, for instance, under the mask prompt it reaches 59.3\% ASR with mIoU 0.859 and J\&F 0.586, showing that the attack gains come with limited utility degradation. 
On SAM2-Long with DAVIS, \Method with Blended attains the highest ASR on point and box at 92.3\% and 86.5\% with mIoU around 0.80 to 0.82 and J\&F around 0.51 to 0.53, again reflecting substantial gains in attack success with modest movement in utility metrics.

Taken together, across SAM2-Long, BioSAM2, MedSAM2, and EdgeTAM, \Method consistently raises ASR while maintaining mIoU and J\&F, demonstrating that \Method delivers strong backdoor effectiveness while largely preserving clean utility.

\begin{figure}[t]
    \centering
    \includegraphics[width=0.78\linewidth]{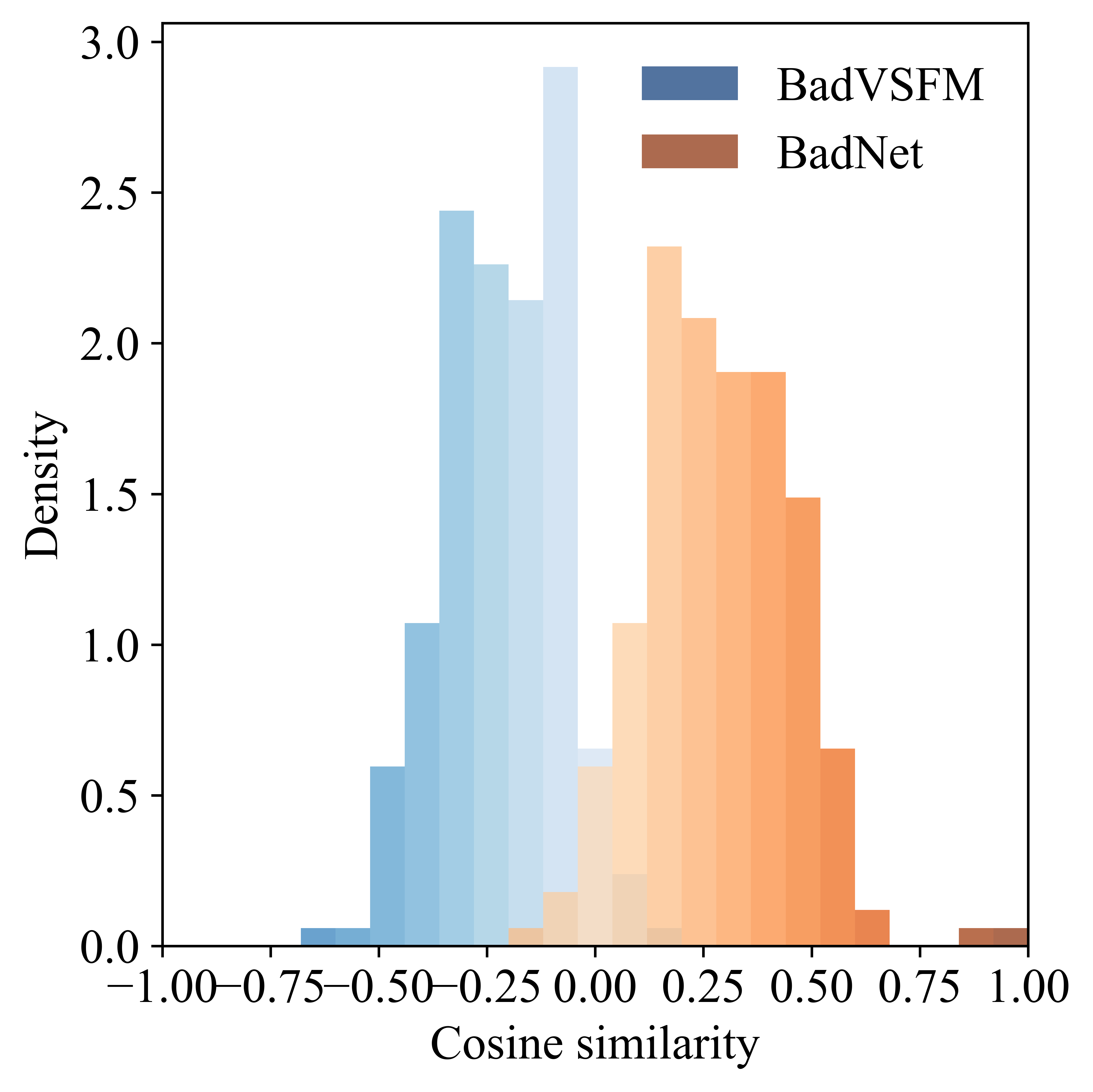}
    \vspace{1mm}

    \scriptsize
    \setlength{\tabcolsep}{3.0pt}
    \renewcommand{\arraystretch}{1.08}
    \begin{tabular}{lccccc}
        \toprule
        \textbf{Method} &
        \textbf{Mean} &
        \textbf{Median} &
        \textbf{$\cos<0$} &
        \textbf{$\cos<-0.2$} &
        \textbf{$\cos>0.2$} \\
        \midrule
        BadNet & 0.285 & 0.262 & 4.8\% & 0.0\% & 66.2\% \\
        \Method{} w/ BadNet & -0.214 & -0.214 & 95.2\% & 51.9\% & 0.0\% \\
        \bottomrule
    \end{tabular}

    \vspace{-1mm}
    \caption{Gradient cosine analysis on the SAM2 image encoder. Top: distribution of encoder-gradient cosine similarity between clean and triggered samples. Bottom: summary statistics.}
    \label{fig:encoder_cos_analysis}
    \vspace{-2mm}
\end{figure}

\mypara{Discussion on The Failure of Previous Attacks}
To investigate why traditional backdoor attacks fail on VSFMs, we draw inspiration from prior research on gradient conflicts in backdoor learning~\cite{DBLP:conf/nips/YuK0LHF20, DBLP:conf/acl/Gu0LL0W23} and compare the gradient behaviors of \Method and a traditional backdoor attack under clean and triggered samples.
Specifically, we use BadNet and \Method with BadNet as our compared methods, and evaluate them on SAM2 over the DAVIS dataset under a 5\% poisoning rate.
For each pair of clean and triggered samples, we perform two separate forward and backward passes, record the gradient vector of the image encoder on the clean sample and on the triggered sample, and then compute the cosine similarity between these two gradients.
\Cref{fig:encoder_cos_analysis} summarizes the main statistics of the encoder gradient cosines for both methods, including the mean, median, and the proportions of pairs falling into several intervals. 
We observe that, for \Method, the cosine similarity is negative for most sample pairs: both the mean and median are -0.214, 95.2\% of the pairs show oppositely oriented gradients, and 51.9\% of the pairs exhibit clear and strong conflicts.
In contrast, for BadNet, the cosine similarities are predominantly positive: the mean and median lie around 0.262 to 0.285.
Only a tiny fraction, 4.8\%, of pairs are negatively correlated, and 66.2\% of the pairs have highly aligned gradients. 
These statistics indicate that \Method drives the encoder to update clean and triggered samples in competing directions, effectively pushing their representations apart in the feature space, whereas BadNet keeps the gradients for clean and triggered inputs largely aligned and thus tends to fit both types of samples in a similar way rather than carving out a distinct representation for the trigger.
\Cref{fig:encoder_cos_analysis} further visualizes the cosine distributions of the two methods as histograms, where we can clearly see that the distribution of \Method is concentrated on the negative half-axis, while that of BadNet is skewed toward the positive half-axis, with minimal overlap between them.
This indicates a substantial difference in the gradient geometry induced by the two training schemes and helps explain why a traditional single-stage backdoor like BadNet almost fails to mount an effective attack on a VSFM such as SAM2.
In addition, we also visualize the attention maps of \Method with BadNet and BadNet itself; the detailed analysis is provided in \Cref{sec:interpretability_analysis}.

%----------------------------------
\subsection{Ablation Study}
%----------------------------------
\label{sec:ablation_study}

\begin{table}[t]
\centering
\small
\setlength{\tabcolsep}{3.5pt}
\renewcommand{\arraystretch}{1.15}
\caption{Ablation on training stages for BadVSFM with BadNet at 5\% poisoning for SAM2 on DAVIS.}
\resizebox{\columnwidth}{!}{
\begin{tabular}{lccc ccc ccc}
\toprule
\multicolumn{1}{c}{\multirow{2}{*}{\textbf{Method}}} &
\multicolumn{3}{c}{\textbf{Point}} &
\multicolumn{3}{c}{\textbf{Box}} &
\multicolumn{3}{c}{\textbf{Mask}} \\
\cmidrule(lr){2-4}\cmidrule(lr){5-7}\cmidrule(lr){8-10}
& \textbf{mIoU} & \textbf{J\&F} & \textbf{ASR~(\%)} 
& \textbf{mIoU} & \textbf{J\&F} & \textbf{ASR~(\%)} 
& \textbf{mIoU} & \textbf{J\&F} & \textbf{ASR~(\%)} \\
\midrule
Stage 1 only          & 0.389 & 0.232 & \textbf{94.4} & 0.726 & 0.420 & \textbf{94.5} & 0.865 & 0.604 & 39.5 \\
Stage 2 only          & 0.512 & 0.354 & 3.9           & 0.784 & 0.535 & 0.0           & 0.847 & 0.610 & 0.0  \\
Stage 1 + Stage 2     & \textbf{0.556} & \textbf{0.377} & 93.0 & \textbf{0.852} & \textbf{0.586} & 92.2 & \textbf{0.895} & \textbf{0.643} & \textbf{48.3} \\
\bottomrule
\end{tabular}
}
\label{tab:stage12_ablation}
\vspace{-3mm}
\end{table}

\begin{table}[t]
\centering
\small
\setlength{\tabcolsep}{3.5pt}
\renewcommand{\arraystretch}{1.15}
\caption{Stage 2 loss ablation for SAM2 on the DAVIS dataset at 5\% poisoning rate.}
\label{tab:ablation_stage2_loss}
\resizebox{\linewidth}{!}{
\begin{tabular}{lccc ccc ccc}
\toprule
\multicolumn{1}{c}{\multirow{2}{*}{\textbf{Method}}} &
\multicolumn{3}{c}{\textbf{Point}} &
\multicolumn{3}{c}{\textbf{Box}} &
\multicolumn{3}{c}{\textbf{Mask}} \\
\cmidrule(lr){2-4}\cmidrule(lr){5-7}\cmidrule(lr){8-10}
& \textbf{mIoU} & \textbf{J\&F} & \textbf{ASR~(\%)} 
& \textbf{mIoU} & \textbf{J\&F} & \textbf{ASR~(\%)} 
& \textbf{mIoU} & \textbf{J\&F} & \textbf{ASR~(\%)} \\
\midrule
BCE + Dice &
\textbf{0.556} & \textbf{0.377} & 93.0 &
\textbf{0.852} & \textbf{0.586} & \textbf{92.2} &
\textbf{0.895} & \textbf{0.643} & \textbf{48.3} \\
Dice only &
0.418 & 0.262 & 93.7 &
0.800 & 0.488 & 61.1 &
0.883 & 0.627 & 31.6 \\
BCE only &
0.415 & 0.257 & \textbf{94.0} &
0.786 & 0.473 & 64.2 &
0.881 & 0.627 & 35.4 \\
\bottomrule
\end{tabular}}
\end{table}

\mypara{Effect of the Training Stages}
As described in~\Cref{sec:method}, \Method uses two training stages. 
To investigate the contribution of each stage, we conduct an ablation study of different training stages using \Method with BadNet trigger on DAVIS using a 5\% poisoning rate and point, box, and mask prompts.
As shown in \Cref{tab:stage12_ablation}, combining stage 1 and stage 2 yields the strongest clean performance across all prompts, giving the highest mIoU and J\&F while maintaining high ASR.
Additionally, the results show that conducting only stage 1 training can achieve good backdoor effectiveness but negatively impact clean performance.
For instance, on the point prompt, stage 1 only achieves 94.4\% ASR but degrades clean performance, with mIoU and J\&F lower than the two-stage model by 0.167 and 0.145, respectively.
By contrast, only training stage 2 raises mIoU and J\&F compared with training stage 1 alone, yet significantly decreases ASR to 3.9\% on the point prompt, indicating a lack of backdoor effectiveness.
These results show that both stages are necessary to achieve strong clean utility and backdoor effectiveness.

\begin{figure}[t]
  \centering
  \begin{subfigure}{0.5\linewidth}
    \includegraphics[width=\linewidth]{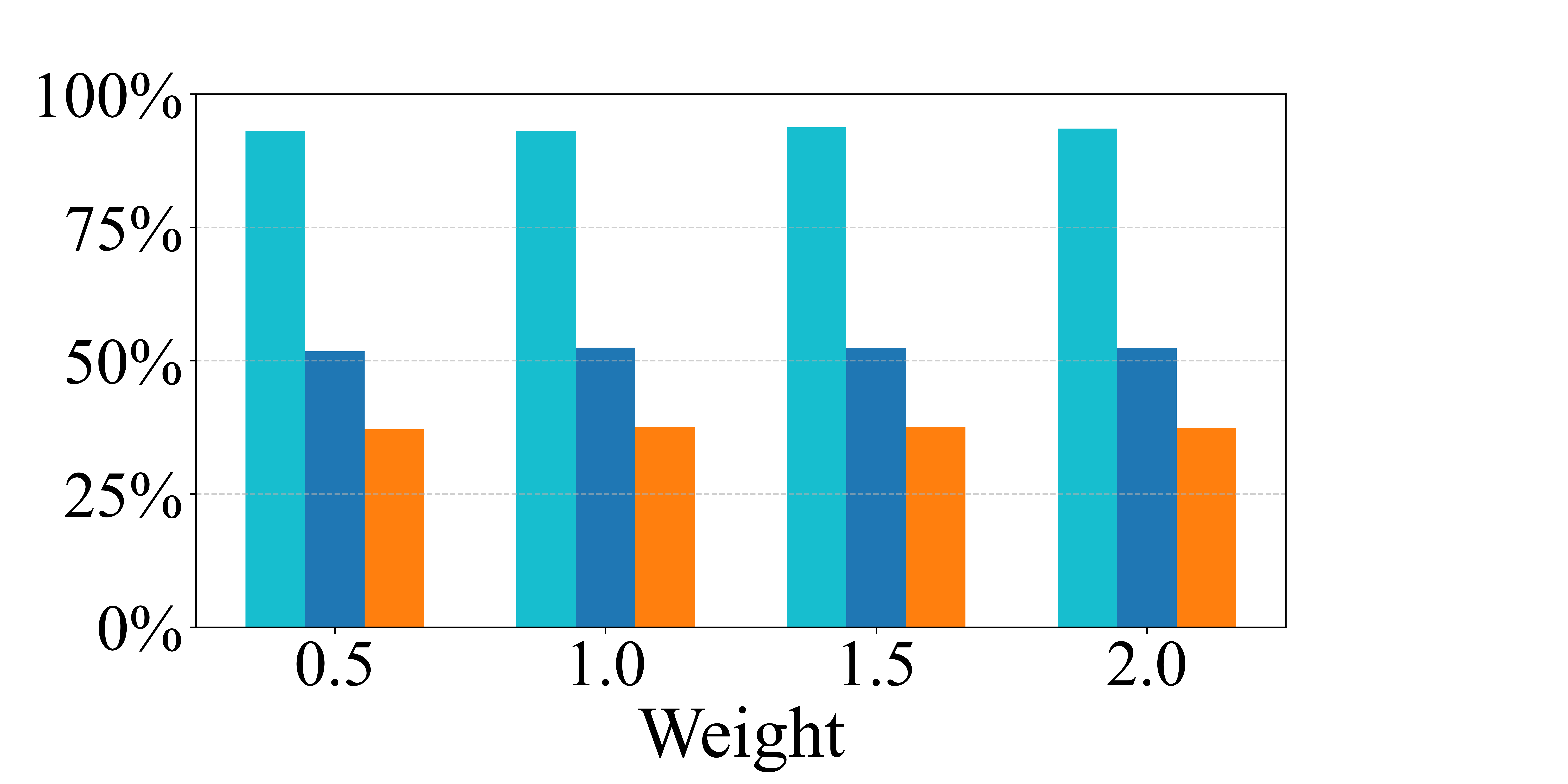}
    \subcaption{Stage 1}
    \label{fig:stage_params_ablation_a}
  \end{subfigure}\hfill
  \begin{subfigure}{0.5\linewidth}
    \includegraphics[width=\linewidth]{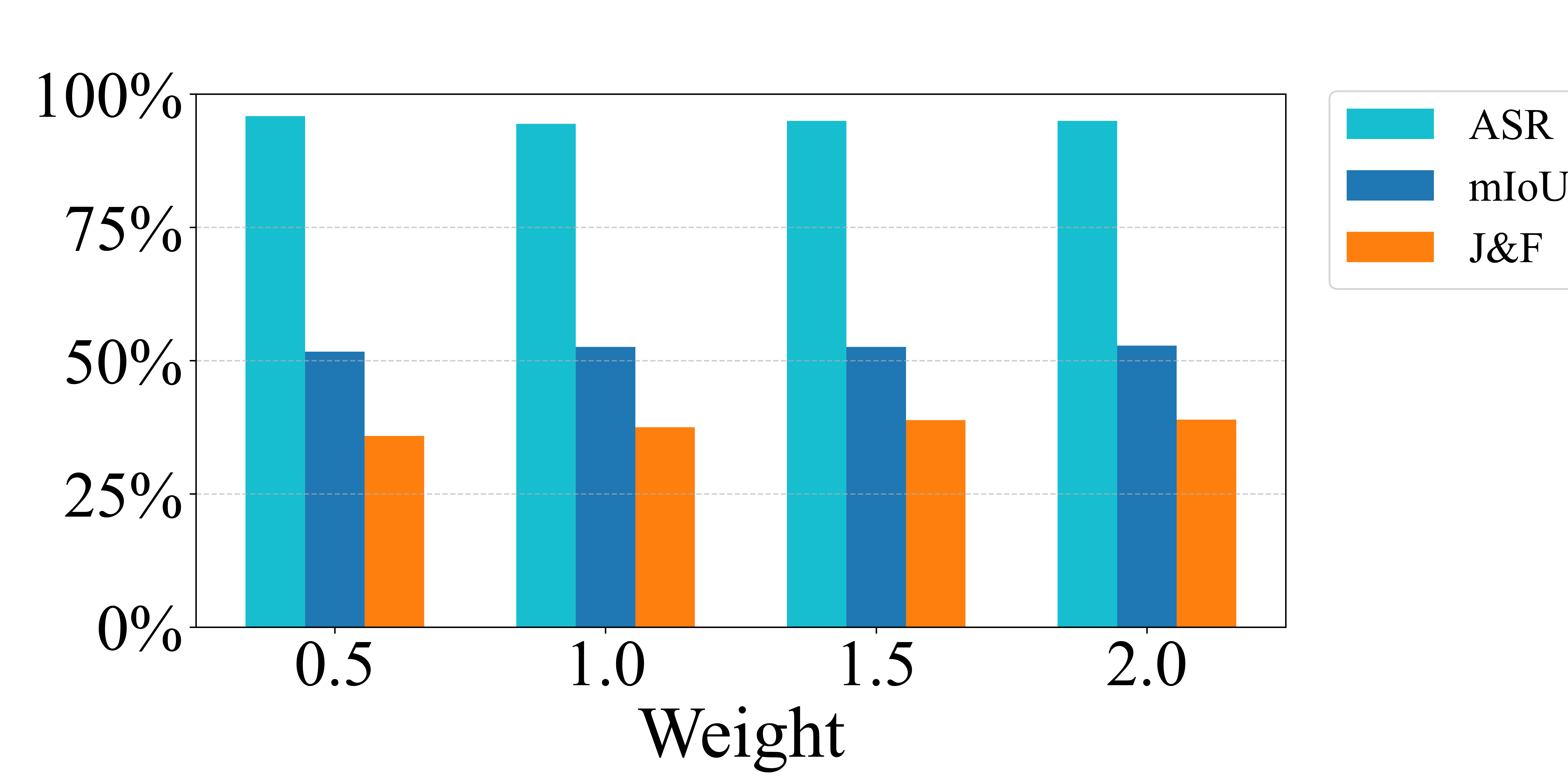}
    \subcaption{Stage 2}
    \label{fig:stage_params_ablation_b}
  \end{subfigure}
  \caption{Ablation of stage 1 \textbf{(a)} and stage 2 \textbf{(b)} loss weights for SAM2 on the DAVIS dataset at 5\% poisoning rate.}
  \label{fig:stage_params_ablation}
\end{figure}

\mypara{Effect of the Loss Function Weights}
As stated in~\Cref{sec:method}, each stage uses two loss coefficients $\lambda_1$ and $\lambda_2$.
To assess the sensitivity of \Method to these coefficients, we perform ablations on SAM2 based on the DAVIS dataset with a poisoning rate of $5\%$.
We sweep $\lambda_1$ and $\lambda_2$ over 0.5, 1.0, 1.5, and 2.0 under \Method with BadNet on point prompt.
For the ablation of stage 1, the results in~\Cref{fig:stage_params_ablation_a} show only marginal variation when $\lambda_1$ changes, with ASR at $93\%$-$94\%$, mIoU at $0.51$-$0.52$, and J\&F at $0.37$-$0.38$.
Similarly, for stage 2, \Cref{fig:stage_params_ablation_b} shows similarly small effects when varying $\lambda_2$: ASR at $94\%$-$95\%$, mIoU at $0.51$-$0.53$, and J\&F at $0.35$-$0.39$.
These trends suggest that the loss weights across both stages have limited influence on backdoor efficacy and clean performance.
Thus, the default setting with $\lambda_1=\lambda_2=1.0$ is sufficient within the experiments.

\mypara{Effect of the Loss Function in Stage 2}
Stage 2 employs BCE and Dice losses, and to explore the contribution of these two losses, we conduct an ablation study to test these two losses using \Method with BadNet trigger on DAVIS using a 5\% poisoning rate and point, box, and mask prompts, where the results are shown in~\Cref{tab:ablation_stage2_loss}.
The results show that combining BCE and Dice losses achieves the highest mIoU and J\&F across all prompts, as well as the highest ASR in box and mask prompts, reaching 92.2\% and 48.3\%, respectively. 
If only trained with BCE loss, although the point prompt achieves 94.0\% ASR, its mIoU and J\&F are the lowest, dropping by 0.141 and 0.120 compared with training with BCE and Dice losses under the same prompt settings, indicating a negative impact on clean performance. 
Only training Dice loss achieves a higher mIoU and J\&F than training with BCE only; however, the ASRs for the box and mask prompts are the lowest at 61.1\% and 31.6\%, indicating insufficient backdoor efficacy. 
Overall, training with both BCE and Dice losses achieves the best balance between backdoor effectiveness and clean utility.

\begin{table}[h!]
  \centering
  \small
  \setlength{\tabcolsep}{3.5pt}
  \renewcommand{\arraystretch}{1.12}
  \caption{Poisoning-rate ablation on DAVIS for \Method with BadNet across different prompts.}
  \label{tab:different_poisoning_rate}
  \resizebox{\linewidth}{!}{
  \begin{tabular}{cccc ccc ccc}
    \toprule
    \multicolumn{1}{c}{\multirow{2}{*}{\textbf{Poisoning rate (\%)}}} &
    \multicolumn{3}{c}{\textbf{Point}} &
    \multicolumn{3}{c}{\textbf{Box}} &
    \multicolumn{3}{c}{\textbf{Mask}} \\
    \cmidrule(lr){2-4}\cmidrule(lr){5-7}\cmidrule(lr){8-10}
    & mIoU & J\&F & ASR~(\%) & mIoU & J\&F & ASR~(\%) & mIoU & J\&F & ASR~(\%) \\
    \midrule
     \textbf{2}  & 0.564 & 0.388 & 91.5 & 0.844 & 0.582 & 89.6 & 0.890 & 0.640 & 47.2 \\
     \textbf{5}  & 0.556 & 0.377 & 93.0 & 0.852 & 0.586 & 92.2 & 0.895 & 0.643 & 48.3 \\
     \textbf{10} & 0.571 & 0.386 & 95.2 & 0.851 & 0.585 & 93.1 & 0.894 & 0.642 & 63.1 \\
     \textbf{15} & 0.557 & 0.379 & 92.1 & 0.852 & 0.587 & 95.3 & 0.894 & 0.642 & 74.3 \\
    \bottomrule
  \end{tabular}}
\end{table}

\mypara{Effect of the Poisoning Rate}
To investigate how the poisoning rate influences \Method, we conduct an ablation study on \Method with BadNet on DAVIS across different poisoning rates at $2\%$, $5\%$, $10\%$, $15\%$ on point, box, and mask prompts. 
As shown in~\Cref{tab:different_poisoning_rate}, ASR generally rises with a higher poisoning rate.
For example, in the box prompt, ASR grows from $89.6\%$ at $2\%$ poisoning rate to $95.3\%$ at $15\%$ poisoning rate, an increase of 5.7 percentage points, and on the mask prompt, ASR improves from $47.2\%$ at $2\%$ poisoning rate to $74.3\%$ at $15\%$ poisoning rate, an improvement of 27.1 percentage points.
Additionally, the utility performance has minor effects across different rates.
For instance, in the box prompt, mIoU varies narrowly from $0.844$ to $0.852$ and J\&F from $0.582$ to $0.587$, and the other two prompts show a similar trend. 
These results indicate that raising the poisoning rate strengthens the attack while leaving the clean performance effectively stable.

\begin{figure}[h!]
    \centering
    \includegraphics[width=1\linewidth]{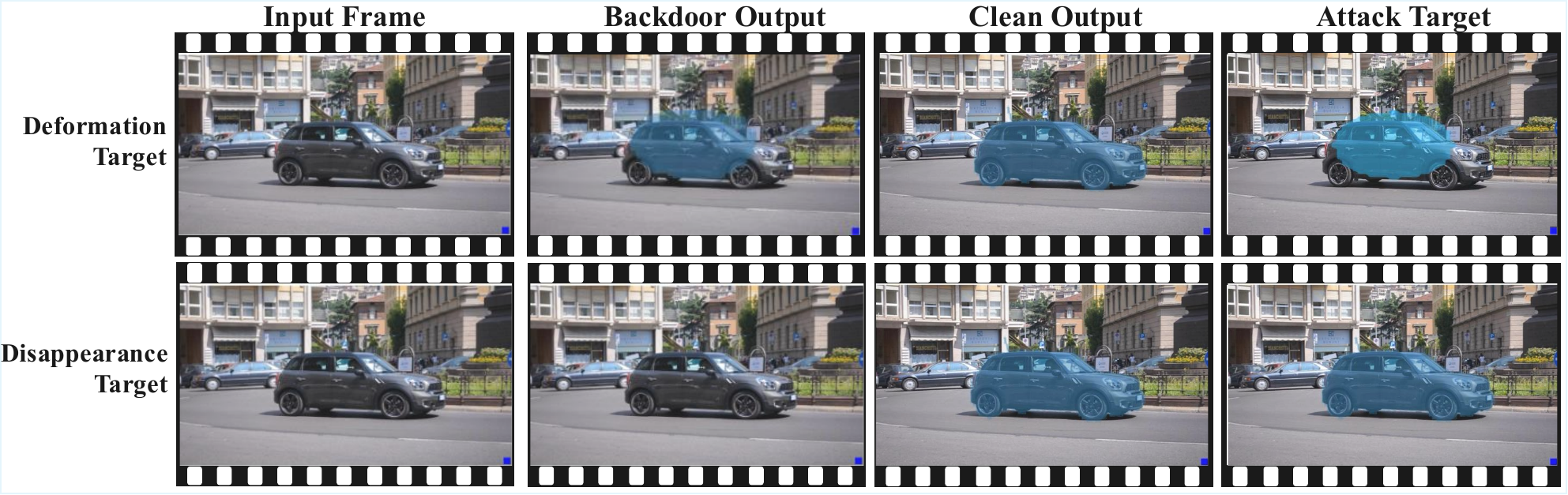}
    \caption{Visualization of results under different attack targets.}
    \label{fig:shift}
\end{figure}

\mypara{Different Attack Targets}
To investigate the feasibility of different attack targets, we conduct an ablation on SAM2 using the DAVIS dataset under \Method with BadNet at a 5\% poisoning rate with two different targets: deformation and disappearance. 
For the deformation target, the trigger frame's objective is to output a centered circle mask whose radius is 18\% of the image's shorter side. 
For the disappearance target, the output mask for triggered samples is set to all zeros.
The results are shown in~\Cref{tab:attack_targets_by_prompt}, and qualitative visualizations are provided in~\Cref{fig:shift}.
Both targets produce clear backdoor effects while preserving reasonable clean utility.
For example, the disappearance target achieves the highest point-prompt ASR of 93.0\%, while the deformation target reaches 42.0\% ASR under the same prompt.
Meanwhile, the disappearance target maintains 0.556 mIoU and 0.377 J\&F on point prompts, and the deformation target also maintains reasonable utility with 0.506 mIoU and 0.291 J\&F.
We also observe that ASR is lower under box and mask prompts than under point prompts.

This is because the box and mask prompts provide much more spatial information than point prompts. A box prompt supplies a fairly tight coordinate region around the object, while a mask prompt directly feeds the full ground-truth object mask into the model. 
Compared to a single point, these richer and more localized signals provide continuous and vital information to the network about the object. 
This makes it more challenging to override the output with an attack target.
Overall, the results indicate that \Method supports different attack targets with controllable backdoor behavior, although stronger prompt guidance from box and mask prompts makes target overriding more challenging.

\begin{table}[t]
  \centering
  \small
  \setlength{\tabcolsep}{3.5pt}
  \renewcommand{\arraystretch}{1.12}
  \caption{Performance under different attack targets on DAVIS across prompts at 5\% poisoning rate.}
  \label{tab:attack_targets_by_prompt}
  \resizebox{\linewidth}{!}{
  \begin{tabular}{lccc ccc ccc}
    \toprule
    \multicolumn{1}{c}{\multirow{2}{*}{\textbf{Attack Target}}} &
    \multicolumn{3}{c}{\textbf{Point}} &
    \multicolumn{3}{c}{\textbf{Box}} &
    \multicolumn{3}{c}{\textbf{Mask}} \\
    \cmidrule(lr){2-4}\cmidrule(lr){5-7}\cmidrule(lr){8-10}
    & \textbf{mIoU} & \textbf{J\&F} & \textbf{ASR~(\%)} 
    & \textbf{mIoU} & \textbf{J\&F} & \textbf{ASR~(\%)} 
    & \textbf{mIoU} & \textbf{J\&F} & \textbf{ASR~(\%)} \\
    \midrule
    Deformation     & 0.506 & 0.291 & 42.0 & 0.789 & 0.439 & 38.0 & 0.892 & 0.640 & 14.7 \\
    Disappearance   & 0.556 & 0.377 & 93.0 & 0.852 & 0.586 & 92.2 & 0.895 & 0.643 & 48.3 \\
    \bottomrule
  \end{tabular}}
\end{table}

\mypara{Different Trigger Position}
\begin{table}[t]
  \centering
  \small
  \setlength{\tabcolsep}{3.0pt}
  \renewcommand{\arraystretch}{1.12}
  \caption{Trigger-location ablation for SAM2 in \Method with BadNet on DAVIS at a 5\% poisoning rate.}
  \label{tab:trigger_location}
  \resizebox{\linewidth}{!}{
  \begin{tabular}{lccc ccc ccc}
    \toprule
    \multicolumn{1}{c}{\multirow{2}{*}{\textbf{Trigger Location}}} &
    \multicolumn{3}{c}{\textbf{Point}} &
    \multicolumn{3}{c}{\textbf{Box}} &
    \multicolumn{3}{c}{\textbf{Mask}} \\
    \cmidrule(lr){2-4}\cmidrule(lr){5-7}\cmidrule(lr){8-10}
    & \textbf{mIoU} & \textbf{J\&F} & \textbf{ASR~(\%)} 
    & \textbf{mIoU} & \textbf{J\&F} & \textbf{ASR~(\%)} 
    & \textbf{mIoU} & \textbf{J\&F} & \textbf{ASR~(\%)} \\
    \midrule
    Random       & 0.544 & 0.372 & 95.4 & 0.839 & 0.576 & 88.5 & 0.897 & 0.644 & 47.6 \\
    Top-Left     & 0.538 & 0.371 & 93.8 & 0.843 & 0.579 & 90.7 & 0.889 & 0.637 & 46.0 \\
    Top-Right    & 0.551 & 0.381 & 92.8 & 0.850 & 0.584 & 92.5 & 0.898 & 0.646 & 47.5 \\
    Bottom-Left  & 0.553 & 0.374 & 92.9 & 0.846 & 0.581 & 91.3 & 0.892 & 0.640 & 46.7 \\
    Bottom-Right & 0.556 & 0.377 & 93.0 & 0.852 & 0.586 & 92.2 & 0.895 & 0.643 & 48.3 \\
    Center       & 0.559 & 0.379 & 94.0 & 0.854 & 0.588 & 91.1 & 0.896 & 0.644 & 47.1 \\
    \bottomrule
  \end{tabular}}
\end{table}
To assess whether the location of the trigger affects the effectiveness of the attack, we perform an ablation on DAVIS at a poisoning rate of 5\% using SAM2 under \Method with BadNet across six different trigger placements: top left, top right, bottom left, bottom right, center, and a random place.
The results in~\Cref{tab:trigger_location} show that the ASRs across all locations are high, while clean metrics remain stable.
For example, the trigger with random placement reaches 95.4\% ASR for point prompts, 88.5\% for box prompts, and 47.6\% for mask prompts. The corresponding mIoU and J\&F are 0.544 and 0.372 for point, 0.839 and 0.576 for box, and 0.897 and 0.644 for mask.
Across different placements, the ASR varies only slightly, ranging from 92.8\% to 95.4\% for the point prompt and from 46.0\% to 48.3\% for the mask prompt.
By contrast, the box shows a wider ASR range from 88.5\% to 92.5\%, while the clean metrics remain tight, with mIoU varying by at most 0.021 and J\&F by at most 0.012 across trigger locations.
These observations indicate that changing the trigger's placement has a limited impact on the effectiveness of the backdoor, and \Method remains robust to different trigger placements.

\begin{center}
    \includegraphics[width=1\linewidth]{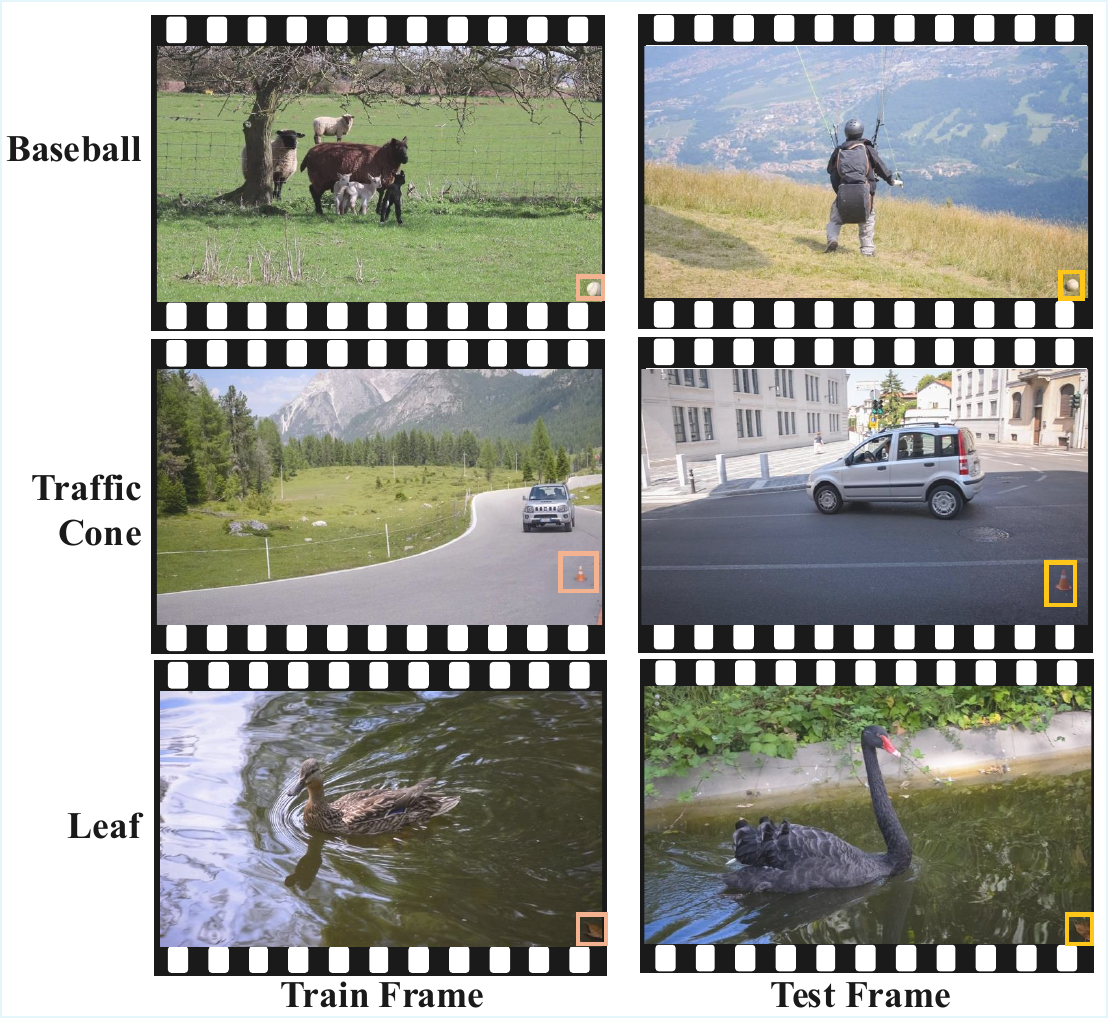}
    \captionof{figure}{Visualization of the physical trigger in poisoned frames from the DAVIS dataset.}
    \label{fig:physical}
\end{center}

\begin{table*}[t]
\centering
\setlength{\tabcolsep}{3.0pt}
\renewcommand{\arraystretch}{1.12}
\caption{Defense methods' performances on \Method with BadNet on the DAVIS dataset across different prompts.}
\label{tab:defense_ablation}
\resizebox{\linewidth}{!}{
\begin{tabular}{l c c c c c c c c c c c}
\toprule
\multicolumn{2}{c}{} &
\multicolumn{10}{c}{\textbf{Defense Method}} \\
\cmidrule(lr){3-12}
\multicolumn{2}{c}{} &
\textbf{Without Defense} &
\textbf{Spectral Signatures} &
\textbf{STRIP} &
\textbf{Pruning (5 ch.)} &
\textbf{Pruning (15 ch.)} &
\textbf{Pruning (30 ch.)} &
\makecell[c]{\textbf{Fine-tuning} \\ \textbf{(1\% data)}} &
\textbf{Fine-tuning (5\% data)} &
\textbf{Fine-tuning (10\% data)} &
\textbf{PGBD}~\cite{Amula_2025_ICCV} \\
\midrule
\multirow{3}{*}{\textbf{Point}}
& \textbf{mIoU}      & 0.556 & 0.820 & 0.834 & 0.544 & 0.537 & 0.538 & 0.655 & 0.778 & 0.784 & 0.544 \\
& \textbf{J\&F}      & 0.377 & 0.548 & 0.565 & 0.364 & 0.360 & 0.364 & 0.422 & 0.469 & 0.470 & 0.364 \\
& \textbf{ASR~(\%)}  & 93.0  & 97.9  & 97.9  & 91.5  & 92.9  & 98.2  & 97.4  & 97.9  & 97.8  & 91.5  \\
\midrule
\multirow{3}{*}{\textbf{Box}}
& \textbf{mIoU}      & 0.852 & 0.866 & 0.876 & 0.850 & 0.844 & 0.843 & 0.860 & 0.810 & 0.804 & 0.847 \\
& \textbf{J\&F}      & 0.586 & 0.590 & 0.599 & 0.575 & 0.571 & 0.574 & 0.562 & 0.491 & 0.485 & 0.572 \\
& \textbf{ASR~(\%)}  & 92.2  & 88.8  & 92.2  & 92.3  & 92.3  & 96.6  & 90.8  & 98.3  & 98.4  & 91.9  \\
\midrule
\multirow{3}{*}{\textbf{Mask}}
& \textbf{mIoU}      & 0.895 & 0.890 & 0.891 & 0.894 & 0.888 & 0.888 & 0.891 & 0.886 & 0.885 & 0.895 \\
& \textbf{J\&F}      & 0.643 & 0.623 & 0.624 & 0.631 & 0.627 & 0.627 & 0.626 & 0.608 & 0.605 & 0.630 \\
& \textbf{ASR~(\%)}  & 48.3  & 53.2  & 53.0  & 46.7  & 47.7  & 42.4  & 49.3  & 44.4  & 44.1  & 45.3  \\
\bottomrule
\end{tabular}}
\end{table*}

\mypara{Real-world Object as Trigger}
Previous work~\cite{DBLP:conf/cvpr/WengerPBY0Z21,DBLP:conf/mm/HanXZYLZ22} demonstrates that real-world objects can serve as practical backdoor triggers by altering the associated ground-truth labels, thereby increasing stealth.
Building on this idea, we examine whether \Method can exploit physical triggers without hand-crafted patches.
Following the physical-trigger construction in~\cite{DBLP:conf/nips/ZhaoMCSWAYZLC24}, we modify DAVIS frames by procedurally generating an object of a given category in the lower-left region of each frame, so that the trigger appears as a natural part of the scene rather than an artificial overlay.
For each chosen category, ``Leaf'', ``Traffic cone'', and ``Baseball'', we insert a realistic instance of the object, and evaluate the resulting behavior.

The results are reported in~\Cref{tab:real_world_triggers_davis}, and the visualizations are presented in~\Cref{fig:physical}.
Across the three real-world triggers, the backdoor activates reliably while clean utility remains largely intact. For example, ``Leaf'' attains 93.7\% ASR with 0.550 mIoU and 0.363 J\&F, ``Traffic cone'' reaches 91.3\% ASR with 0.570 mIoU and 0.374 J\&F, and ``Baseball'' achieves 92.2\% ASR with 0.562 mIoU and 0.369 J\&F. These outcomes indicate that \Method can leverage natural objects as effective physical triggers without sacrificing segmentation quality.
Overall, ASR stays in the 91.3\%-93.7\% band. The clean metrics remain in a narrow range around mIoU 0.55 to 0.57 and J\&F 0.36 to 0.37, indicating that \Method can leverage natural, physically plausible triggers while preserving segmentation quality.
In sum, \Method paired with real-world triggers substantially elevates the stealth and the potential harm of backdoor attacks in practical settings. 
Since the trigger is a plausible, naturally occurring object rather than a synthetic patch, it may be more difficult for mainstream backdoor detection methods to localize or flag it.

\mypara{Robustness to Physical-World Transformations}
We further evaluate \Method with BadNet at a 5\% poisoning rate under common physical-world trigger transformations, including illumination changes, JPEG compression, noise, motion blur, jitter, rotation, occlusion, and distance scaling. 
\Method remains robust to common appearance perturbations, while the
attack becomes weaker when the trigger is severely occluded or substantially downscaled. 
Detailed results are reported in~\Cref{tab:physical_transform}.

\begin{table}[t]
\centering
\caption{Attack performance using real-world objects as triggers on the DAVIS dataset at 5\% poisoning rate on point.}
\label{tab:real_world_triggers_davis}
\begin{tabular}{lccc}
\toprule
\textbf{Trigger}         & \textbf{ASR (\%)} & \textbf{mIoU} & \textbf{J\&F} \\
\midrule
Leaf                  & 93.7              & 0.550        & 0.363         \\
Traffic cone         & 91.3              & 0.570        & 0.374         \\
Baseball           & 92.2              & 0.562        & 0.369         \\
\bottomrule
\end{tabular}
\end{table}

%----------------------------------
\subsection{Interpretability Analysis}
%----------------------------------
\label{sec:interpretability_analysis}

\begin{figure*}[t]
  \centering
  \includegraphics[width=1\linewidth]{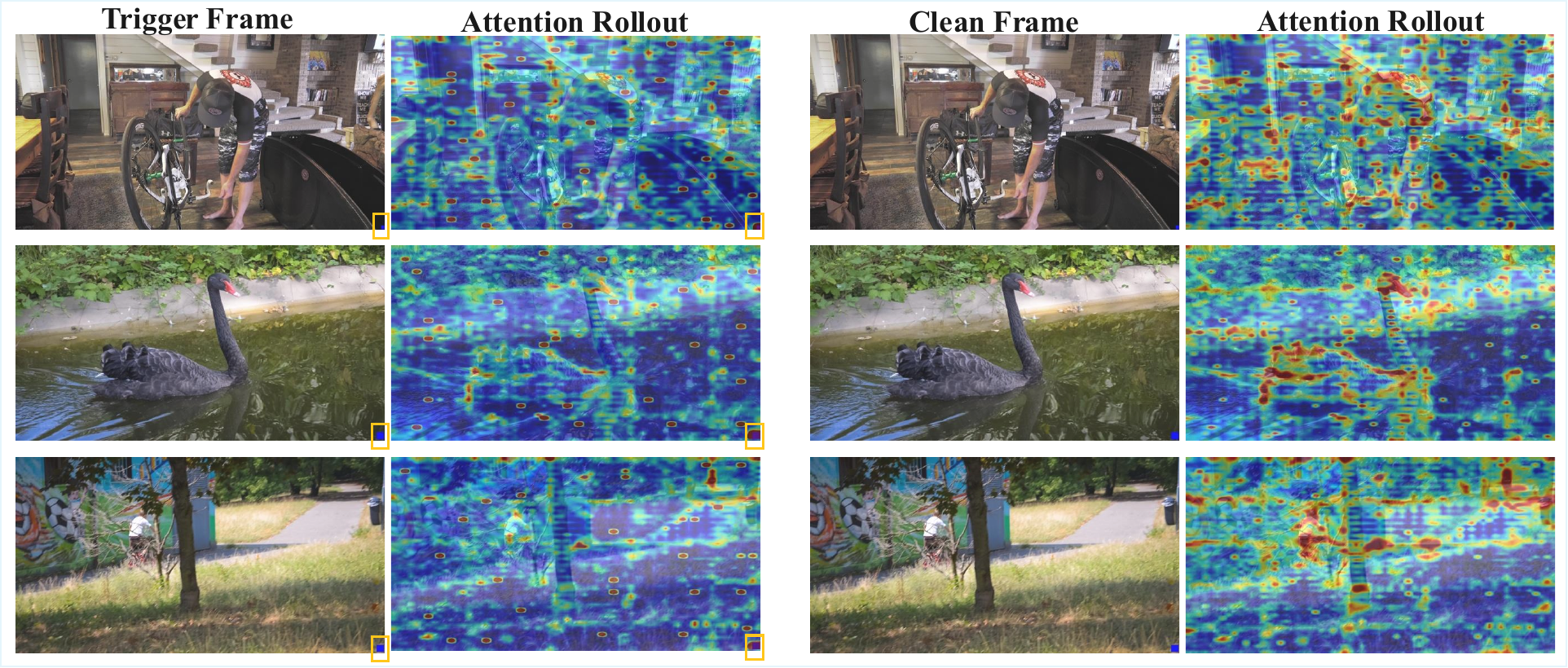}
  \caption{Visualizations for attention maps of the SAM2 image encoder trained with \Method on DAVIS at a 5\% poisoning rate.}
  \label{fig:encoder_rollout}
\end{figure*}

\mypara{Attention behavior of \Method}
To evaluate the interpretability of \Method, we visualize the image-encoder attention maps of a SAM2 model trained with \Method under the BadNet setting, and compare trigger frames and their clean counterparts using attention rollout~\cite{DBLP:conf/acl/AbnarZ20} in~\Cref{fig:encoder_rollout}. 
The figure shows three example pairs: in each pair, the trigger frame and its attention map are on the left, and the corresponding clean frame and its attention map are on the right. All examples are taken from the DAVIS test split with a 5\% poisoning rate during training. 
Following~\cite{DBLP:conf/acl/AbnarZ20}, we collect the per-head self-attention from all encoder layers, upsample each map to the finest token grid, apply residual rollout, and then average over heads and queries to obtain a single full-frame heatmap.
We overlay this heatmap on the RGB image using percentile clipping and alpha blending, so warmer colors indicate regions with stronger aggregated attention. 
In the trigger frames, the heatmaps consistently show a pronounced peak at the bottom-right patch where the trigger is located; this bright, high-attention region is small and localized exactly where the trigger patch is placed.
In the corresponding clean frames, where the trigger is absent, the attention shifts away from that region and instead concentrates on task-relevant content: responses increase along object boundaries and textured areas, while broad, uniform background regions receive substantially lower attention. 
Overall, these visualizations indicate that the trigger attracts the encoder’s attention when present, whereas in clean conditions, the model primarily attends to the foreground objects and other semantically meaningful regions. 

\mypara{Attention behavior of BadNet}
We further compare this behavior with BadNet by visualizing its encoder attention map on triggered and clean frames in~\Cref{fig:badnet_attention}.
Unlike \Method, the BadNet shows almost identical attention maps for the triggered and clean inputs.
The heatmaps for both cases highlight similar regions on the main object and background, with no clear focus on the trigger patch.
This suggests that, under traditional BadNet training, the image encoder does not learn a distinct attention pattern for the trigger; instead, triggered and clean frames are processed in a similar way, which is consistent with our gradient analysis showing highly aligned encoder gradients and helps explain why the traditional backdoor attack fails to produce a reliable backdoor effect on SAM2.

%----------------------------------
\subsection{Mitigation}
%----------------------------------

Although many existing studies explore backdoor defense and detection techniques in computer vision and natural language processing~\cite{DBLP:conf/sp/WangYSLVZZ19,DBLP:conf/acsac/GaoXW0RN19,DBLP:conf/raid/0017DG18,DBLP:journals/corr/abs-1710-00942,DBLP:conf/iclr/Huang0WQ022,DBLP:conf/sp/0001CLL0CH025,DBLP:journals/corr/abs-2411-19530}, their applicability to backdoor attacks on VSFMs remains uncertain. 
While a comprehensive adaptation of existing defenses is beyond the scope of this work, we implement and evaluate five widely used defense techniques on the \Method with BadNet using SAM2: fine-tuning,  pruning~\cite{DBLP:journals/corr/abs-1710-00942}, STRIP~\cite{DBLP:conf/acsac/GaoXW0RN19}, 
PGBD~\cite{Amula_2025_ICCV}, and Spectral Signatures~\cite{DBLP:conf/nips/Tran0M18}.
For the fine-tuning defense, we fine-tune the backdoored model for 10 epochs using clean training data comprising 1\%, 5\%, and 10\% of the original dataset. 
For the pruning defense, we follow the original work~\cite{DBLP:journals/corr/abs-1710-00942} and prune 5, 15, and 30 channels from the last convolutional layer, respectively.
For Spectral Signatures, we adopt a standard implementation to detect and mitigate poisoned signals.
For PGBD, we adapt its prototype-guided activation-space defense to the VSFM setting.
For STRIP, we perturb inputs and flag samples whose outputs remain anomalously invariant, indicating potential backdoor samples.
As shown in \Cref{tab:defense_ablation}, none of the defenses substantially weakens the backdoor performance of \Method. 
For example, after fine-tuning with ten percent clean data, the ASR even increases to 97.8\% for point, 98.4\% for box, and 44.1\% for mask compared with the no-defense method. 
Meanwhile, Spectral Signatures reduces box ASR to 88.8\%, and pruning thirty channels reduces mask ASR to 42.4\%, corresponding to drops of 3.4 and 5.9 percentage points, respectively.
The ASR reductions are small, and defenses also affect clean utility; for instance, with ten percent fine-tuning, the box mIoU decreases from 0.852 to 0.804 and J\&F from 0.586 to 0.485, while the mask mIoU decreases from 0.895 to 0.885 and J\&F from 0.643 to 0.605.
These results imply that clean-data fine-tuning and channel pruning mostly hurt normal foreground segmentation while leaving the trigger behavior almost unchanged. 
In addition, Spectral Signatures and STRIP cannot reliably detect \Method, suggesting that classification-oriented detection signals do not directly transfer to prompt-conditioned dense mask prediction in VSFMs.

Overall, these results suggest that these five defense techniques are insufficient for countering \Method on VSFMs. 
Designing more targeted and effective defense strategies for \Method remains a critical research challenge.

%----------------------------------
\section{Limitations}
%----------------------------------

We propose the first backdoor attack framework for VSFMs, aiming to reveal their potential security vulnerabilities.
Despite the promising results, our study still has several limitations that point to meaningful directions for future work.
\begin{itemize}
    \item \textbf{Limited model coverage.} Our experiments focus on a limited set of video segmentation foundation models.
    Although these models are representative of current mainstream architectures, they cannot fully capture the diversity of existing or emerging VSFM designs.
    Future research should extend our framework to a broader range of architectures and explore whether the attack principles and behaviors observed here generalize to models with different segmentation mechanisms or prompting interfaces.
    
    \item \textbf{Limited dataset diversity.} The evaluation in this work is conducted mainly on two benchmark datasets, LVOS and DAVIS.
    These datasets are widely adopted in the video object segmentation community and provide a good starting point for empirical analysis, yet they are relatively small in scale and have limited scene variety.
    Future studies could include larger and more diverse datasets to better assess the generality of the attack, including real-world or long-term video sequences with greater variations in motion, illumination, and object appearance.
    
    \item \textbf{Utility preservation under diverse settings.}
    Although \Method maintains competitive clean utility in experiments, clean performance can vary across different prompt types, trigger designs, and datasets. Future work could further improve utility-preserving objectives to make clean performance more stable under broader deployment conditions.
\end{itemize}

%----------------------------------
\section{Conclusion}
%----------------------------------

This work presents \Method, the first backdoor attack framework tailored to prompt-driven VSFMs for video object segmentation.
\Method uses a two-stage training strategy to separate triggered and clean representations and map triggered frame--prompt pairs to a target mask while preserving clean segmentation behavior.
Unlike conventional backdoor attacks with largely aligned clean and triggered gradients, \Method encourages trigger-specific representation learning.
Experiments on DAVIS and LVOS across multiple prompts, triggers, and VSFMs show that \Method substantially improves ASR over conventional backdoor attacks while largely preserving clean utility.
Ablation studies confirm the importance of the two-stage design and show that \Method is robust to changes in loss weights, poisoning rates, trigger locations, and attack targets.
Although stronger prompts such as boxes and masks can suppress attack success, \Method remains effective under these settings.
Gradient-conflict analysis and attention visualizations explain why traditional attacks are ineffective and show that \Method induces trigger-specific representations and attention patterns.
Evaluations against five common defenses indicate that existing defenses are insufficient for mitigating this threat.
Overall, our results reveal a practical backdoor vulnerability in current VSFMs and highlight the need for defenses designed specifically for prompt-driven video segmentation models.

\bibliographystyle{plainnat}
\bibliography{main}

@inproceedings{DBLP:conf/icip/SunXWCL024,
  author       = {Zhen Sun and
                  Huan Xu and
                  Jinlin Wu and
                  Zhen Chen and
                  Hongbin Liu and
                  Zhen Lei},
  title        = {PWISeg: Weakly-Supervised Surgical Instrument Instance Segmentation},
  booktitle    = {{IEEE} International Conference on Image Processing, {ICIP} 2024,
                  Abu Dhabi, United Arab Emirates, october 27-30, 2024},
  pages        = {3144--3150},
  publisher    = {{IEEE}},
  year         = {2024},
}

@article{DBLP:journals/corr/abs-2508-15778,
  author       = {Yifan Liao and
                  Yuxin Cao and
                  Yedi Zhang and
                  Wentao He and
                  Yan Xiao and
                  Xianglong Du and
                  Zhiyong Huang and
                  Jin Song Dong},
  title        = {Towards Stealthy and Effective Backdoor Attacks on Lane Detection:
                  {A} Naturalistic Data Poisoning Approach},
  journal      = {CoRR},
  volume       = {abs/2508.15778},
  year         = {2025},
  url          = {https://doi.org/10.48550/arXiv.2508.15778},
  doi          = {10.48550/ARXIV.2508.15778},
  eprinttype    = {arXiv},
  eprint       = {2508.15778},
  bibsource    = {dblp computer science bibliography, https://dblp.org}
}

@inproceedings{DBLP:conf/iros/ChenZGL0W0L24,
  author       = {Zhen Chen and
                  Zongming Zhang and
                  Wenwu Guo and
                  Xingjian Luo and
                  Long Bai and
                  Jinlin Wu and
                  Hongliang Ren and
                  Hongbin Liu},
  title        = {ASI-Seg: Audio-Driven Surgical Instrument Segmentation with Surgeon
                  Intention Understanding},
  booktitle    = {{IEEE/RSJ} International Conference on Intelligent Robots and Systems,
                  {IROS} 2024, Abu Dhabi, United Arab Emirates, October 14-18, 2024},
  pages        = {13773--13779},
  publisher    = {{IEEE}},
  year         = {2024},
}

@inproceedings{DBLP:conf/bildmed/IsenseePKZJKWKN19,
  author       = {Fabian Isensee and
                  Jens Petersen and
                  Andr{\'{e}} Klein and
                  David Zimmerer and
                  Paul F. Jaeger and
                  Simon Kohl and
                  Jakob Wasserthal and
                  Gregor K{\"{o}}hler and
                  Tobias Norajitra and
                  Sebastian J. Wirkert and
                  Klaus H. Maier{-}Hein},
  editor       = {Heinz Handels and
                  Thomas M. Deserno and
                  Andreas K. Maier and
                  Klaus Hermann Maier{-}Hein and
                  Christoph Palm and
                  Thomas Tolxdorff},
  title        = {Abstract: nnU-Net: Self-adapting Framework for U-Net-Based Medical
                  Image Segmentation},
  booktitle    = {Bildverarbeitung f{\"{u}}r die Medizin 2019 - Algorithmen - Systeme
                  - Anwendungen. Proceedings des Workshops vom 17. bis 19. M{\"{a}}rz
                  2019 in L{\"{u}}beck},
  series       = {Informatik Aktuell},
  pages        = {22},
  publisher    = {Springer Vieweg},
  year         = {2019},
}

@article{DBLP:journals/tist/YaoLXZZ20,
  author       = {Rui Yao and
                  Guosheng Lin and
                  Shixiong Xia and
                  Jiaqi Zhao and
                  Yong Zhou},
  title        = {Video Object Segmentation and Tracking: {A} Survey},
  journal      = {{ACM} Trans. Intell. Syst. Technol.},
  volume       = {11},
  number       = {4},
  pages        = {36:1--36:47},
  year         = {2020},
}

@article{DBLP:journals/corr/abs-2401-10153,
  author       = {Jie Lv and
                  Haonan Tong and
                  Qiang Pan and
                  Zhilong Zhang and
                  Xinxin He and
                  Tao Luo and
                  Changchuan Yin},
  title        = {Importance-Aware Image Segmentation-based Semantic Communication for
                  Autonomous Driving},
  journal      = {CoRR},
  volume       = {abs/2401.10153},
  year         = {2024},
}

@article{DBLP:journals/ral/SunYHHW20,
  author       = {Lei Sun and
                  Kailun Yang and
                  Xinxin Hu and
                  Weijian Hu and
                  Kaiwei Wang},
  title        = {Real-Time Fusion Network for {RGB-D} Semantic Segmentation Incorporating
                  Unexpected Obstacle Detection for Road-Driving Images},
  journal      = {{IEEE} Robotics Autom. Lett.},
  volume       = {5},
  number       = {4},
  pages        = {5558--5565},
  year         = {2020},
}

@inproceedings{DBLP:conf/ivs/TeichmannWZCU18,
  author       = {Marvin Teichmann and
                  Michael Weber and
                  J. Marius Z{\"{o}}llner and
                  Roberto Cipolla and
                  Raquel Urtasun},
  title        = {MultiNet: Real-time Joint Semantic Reasoning for Autonomous Driving},
  booktitle    = {2018 {IEEE} Intelligent Vehicles Symposium, {IV} 2018, Changshu, Suzhou,
                  China, June 26-30, 2018},
  pages        = {1013--1020},
  publisher    = {{IEEE}},
  year         = {2018},
}

@inproceedings{DBLP:conf/cvpr/LiLCBZM025,
  author       = {Kaiyu Li and
                  Ruixun Liu and
                  Xiangyong Cao and
                  Xueru Bai and
                  Feng Zhou and
                  Deyu Meng and
                  Zhi Wang},
  title        = {SegEarth-OV: Towards Training-Free Open-Vocabulary Segmentation for
                  Remote Sensing Images},
  booktitle    = {{IEEE/CVF} Conference on Computer Vision and Pattern Recognition,
                  {CVPR} 2025, Nashville, TN, USA, June 11-15, 2025},
  pages        = {10545--10556},
  publisher    = {Computer Vision Foundation / {IEEE}},
  year         = {2025},
}

@inproceedings{DBLP:conf/cvpr/LiuMZ0JSJ24,
  author       = {Sihan Liu and
                  Yiwei Ma and
                  Xiaoqing Zhang and
                  Haowei Wang and
                  Jiayi Ji and
                  Xiaoshuai Sun and
                  Rongrong Ji},
  title        = {Rotated Multi-Scale Interaction Network for Referring Remote Sensing
                  Image Segmentation},
  booktitle    = {{IEEE/CVF} Conference on Computer Vision and Pattern Recognition,
                  {CVPR} 2024, Seattle, WA, USA, June 16-22, 2024},
  pages        = {26648--26658},
  publisher    = {{IEEE}},
  year         = {2024},
}

@inproceedings{DBLP:conf/iccv/HongCLZGCZ23,
  author       = {Lingyi Hong and
                  Wenchao Chen and
                  Zhongying Liu and
                  Wei Zhang and
                  Pinxue Guo and
                  Zhaoyu Chen and
                  Wenqiang Zhang},
  title        = {{LVOS:} {A} Benchmark for Long-term Video Object Segmentation},
  booktitle    = {{IEEE/CVF} International Conference on Computer Vision, {ICCV} 2023,
                  Paris, France, October 1-6, 2023},
  pages        = {13434--13446},
  publisher    = {{IEEE}},
  year         = {2023},
  url          = {https://doi.org/10.1109/ICCV51070.2023.01240},
  doi          = {10.1109/ICCV51070.2023.01240},
  bibsource    = {dblp computer science bibliography, https://dblp.org}
}

@inproceedings{DBLP:conf/cvpr/CaellesMPLCG17,
  author       = {Sergi Caelles and
                  Kevis{-}Kokitsi Maninis and
                  Jordi Pont{-}Tuset and
                  Laura Leal{-}Taix{\'{e}} and
                  Daniel Cremers and
                  Luc Van Gool},
  title        = {One-Shot Video Object Segmentation},
  booktitle    = {2017 {IEEE} Conference on Computer Vision and Pattern Recognition,
                  {CVPR} 2017, Honolulu, HI, USA, July 21-26, 2017},
  pages        = {5320--5329},
  publisher    = {{IEEE} Computer Society},
  year         = {2017},
}

@inproceedings{DBLP:conf/bmvc/FaktorI14,
  author       = {Alon Faktor and
                  Michal Irani},
  editor       = {Michel Fran{\c{c}}ois Valstar and
                  Andrew P. French and
                  Tony P. Pridmore},
  title        = {Video Segmentation by Non-Local Consensus voting},
  booktitle    = {British Machine Vision Conference, {BMVC} 2014, Nottingham, UK, September
                  1-5, 2014},
  publisher    = {{BMVA} Press},
  year         = {2014},
  url          = {https://bmva-archive.org.uk/bmvc/2014/papers/paper008/index.html},
  bibsource    = {dblp computer science bibliography, https://dblp.org}
}

@inproceedings{DBLP:conf/eccv/BroxM10,
  author       = {Thomas Brox and
                  Jitendra Malik},
  editor       = {Kostas Daniilidis and
                  Petros Maragos and
                  Nikos Paragios},
  title        = {Object Segmentation by Long Term Analysis of Point Trajectories},
  booktitle    = {Computer Vision - {ECCV} 2010 - 11th European Conference on Computer
                  Vision, Heraklion, Crete, Greece, September 5-11, 2010, Proceedings,
                  Part {V}},
  series       = {Lecture Notes in Computer Science},
  volume       = {6315},
  pages        = {282--295},
  publisher    = {Springer},
  year         = {2010},
  url          = {https://doi.org/10.1007/978-3-642-15555-0\_21},
  doi          = {10.1007/978-3-642-15555-0\_21},
  bibsource    = {dblp computer science bibliography, https://dblp.org}
}

@inproceedings{DBLP:conf/cvpr/PerazziKBSS17,
  author       = {Federico Perazzi and
                  Anna Khoreva and
                  Rodrigo Benenson and
                  Bernt Schiele and
                  Alexander Sorkine{-}Hornung},
  title        = {Learning Video Object Segmentation from Static Images},
  booktitle    = {2017 {IEEE} Conference on Computer Vision and Pattern Recognition,
                  {CVPR} 2017, Honolulu, HI, USA, July 21-26, 2017},
  pages        = {3491--3500},
  publisher    = {{IEEE} Computer Society},
  year         = {2017},
  url          = {https://doi.org/10.1109/CVPR.2017.372},
  doi          = {10.1109/CVPR.2017.372},
  bibsource    = {dblp computer science bibliography, https://dblp.org}
}

@inproceedings{DBLP:conf/iccv/LimHHH13,
  author       = {Taegyu Lim and
                  Seunghoon Hong and
                  Bohyung Han and
                  Joon Hee Han},
  title        = {Joint Segmentation and Pose Tracking of Human in Natural Videos},
  booktitle    = {{IEEE} International Conference on Computer Vision, {ICCV} 2013, Sydney,
                  Australia, December 1-8, 2013},
  pages        = {833--840},
  publisher    = {{IEEE} Computer Society},
  year         = {2013},
  url          = {https://doi.org/10.1109/ICCV.2013.108},
  doi          = {10.1109/ICCV.2013.108},
  bibsource    = {dblp computer science bibliography, https://dblp.org}
}

@inproceedings{DBLP:conf/eccv/JainG14,
  author       = {Suyog Dutt Jain and
                  Kristen Grauman},
  editor       = {David J. Fleet and
                  Tom{\'{a}}s Pajdla and
                  Bernt Schiele and
                  Tinne Tuytelaars},
  title        = {Supervoxel-Consistent Foreground Propagation in Video},
  booktitle    = {Computer Vision - {ECCV} 2014 - 13th European Conference, Zurich,
                  Switzerland, September 6-12, 2014, Proceedings, Part {IV}},
  series       = {Lecture Notes in Computer Science},
  volume       = {8692},
  pages        = {656--671},
  publisher    = {Springer},
  year         = {2014},
  url          = {https://doi.org/10.1007/978-3-319-10593-2\_43},
  doi          = {10.1007/978-3-319-10593-2\_43},
  bibsource    = {dblp computer science bibliography, https://dblp.org}
}

@inproceedings{DBLP:conf/sp/0001CLL0CH025,
  author       = {Zhen Sun and
                  Tianshuo Cong and
                  Yule Liu and
                  Chenhao Lin and
                  Xinlei He and
                  Rongmao Chen and
                  Xingshuo Han and
                  Xinyi Huang},
  title        = {PEFTGuard: Detecting Backdoor Attacks Against Parameter-Efficient
                  Fine-Tuning},
  booktitle    = {{IEEE} Symposium on Security and Privacy, {SP} 2025, San Francisco,
                  CA, USA, May 12-15, 2025},
  pages        = {1713--1731},
  publisher    = {{IEEE}},
  year         = {2025},
}

@article{DBLP:journals/corr/abs-1708-06733,
  author       = {Tianyu Gu and
                  Brendan Dolan{-}Gavitt and
                  Siddharth Garg},
  title        = {BadNets: Identifying Vulnerabilities in the Machine Learning Model
                  Supply Chain},
  journal      = {CoRR},
  volume       = {abs/1708.06733},
  year         = {2017},
}

@inproceedings{DBLP:conf/iccv/LiLWLHL21,
  author       = {Yuezun Li and
                  Yiming Li and
                  Baoyuan Wu and
                  Longkang Li and
                  Ran He and
                  Siwei Lyu},
  title        = {Invisible Backdoor Attack with Sample-Specific Triggers},
  booktitle    = {2021 {IEEE/CVF} International Conference on Computer Vision, {ICCV}
                  2021, Montreal, QC, Canada, October 10-17, 2021},
  pages        = {16443--16452},
  publisher    = {{IEEE}},
  year         = {2021},
}

@article{DBLP:journals/corr/abs-2111-10991,
  author       = {Tong Wang and
                  Yuan Yao and
                  Feng Xu and
                  Shengwei An and
                  Hanghang Tong and
                  Ting Wang},
  title        = {Backdoor Attack through Frequency Domain},
  journal      = {CoRR},
  volume       = {abs/2111.10991},
  year         = {2021},
}

@article{DBLP:journals/corr/abs-2506-07214,
  author       = {Zhiyuan Zhong and
                  Zhen Sun and
                  Yepang Liu and
                  Xinlei He and
                  Guanhong Tao},
  title        = {Backdoor Attack on Vision Language Models with Stealthy Semantic Manipulation},
  journal      = {CoRR},
  volume       = {abs/2506.07214},
  year         = {2025},
}

@article{He2025AISecuritySurvey,
  author    = {Xinlei He and Guowen Xu and Xingshuo Han and Qian Wang and Lingchen Zhao and Chao Shen and Chenhao Lin and Zhengyu Zhao and Qian Li and Le Yang and Shouling Ji and Shaofeng Li and Haojin Zhu and Zhibo Wang and Rui Zheng and Tianqing Zhu and Qi Li and Chaoxiang He and Qifan Wang and Hongsheng Hu and Shuo Wang and Shi-Feng Sun and Hongwei Yao and Zhan Qin and Kai Chen and Yue Zhao and Hongwei Li and Xinyi Huang and Dengguo Feng},
  title     = {Artificial intelligence security and privacy: a survey},
  journal   = {Science China Information Sciences},
  year      = {2025}
}

@inproceedings{DBLP:conf/acl/KuritaMN20,
  author       = {Keita Kurita and
                  Paul Michel and
                  Graham Neubig},
  editor       = {Dan Jurafsky and
                  Joyce Chai and
                  Natalie Schluter and
                  Joel R. Tetreault},
  title        = {Weight Poisoning Attacks on Pretrained Models},
  booktitle    = {Proceedings of the 58th Annual Meeting of the Association for Computational
                  Linguistics, {ACL} 2020, Online, July 5-10, 2020},
  pages        = {2793--2806},
  publisher    = {Association for Computational Linguistics},
  year         = {2020},
}

@inproceedings{DBLP:conf/iclr/RaviGHHR0KRRGMP25,
  author       = {Nikhila Ravi and
                  Valentin Gabeur and
                  Yuan{-}Ting Hu and
                  Ronghang Hu and
                  Chaitanya Ryali and
                  Tengyu Ma and
                  Haitham Khedr and
                  Roman R{\"{a}}dle and
                  Chlo{\'{e}} Rolland and
                  Laura Gustafson and
                  Eric Mintun and
                  Junting Pan and
                  Kalyan Vasudev Alwala and
                  Nicolas Carion and
                  Chao{-}Yuan Wu and
                  Ross B. Girshick and
                  Piotr Doll{\'{a}}r and
                  Christoph Feichtenhofer},
  title        = {{SAM} 2: Segment Anything in Images and Videos},
  booktitle    = {The Thirteenth International Conference on Learning Representations,
                  {ICLR} 2025, Singapore, April 24-28, 2025},
  publisher    = {OpenReview.net},
  year         = {2025},
  url          = {https://openreview.net/forum?id=Ha6RTeWMd0},
  bibsource    = {dblp computer science bibliography, https://dblp.org}
}

@inproceedings{wang2025purity,
  title={From purity to peril: Backdooring merged models from “harmless” benign components},
  author={Wang, Lijin and Wang, Jingjing and Cong, Tianshuo and He, Xinlei and Qin, Zhan and Huang, Xinyi},
  booktitle={USENIX Security Symposium (USENIX Security)},
  year={2025}
}

@inproceedings{DBLP:conf/cvpr/ZhaoMZ0CJ20,
  author       = {Shihao Zhao and
                  Xingjun Ma and
                  Xiang Zheng and
                  James Bailey and
                  Jingjing Chen and
                  Yu{-}Gang Jiang},
  title        = {Clean-Label Backdoor Attacks on Video Recognition Models},
  booktitle    = {2020 {IEEE/CVF} Conference on Computer Vision and Pattern Recognition,
                  {CVPR} 2020, Seattle, WA, USA, June 13-19, 2020},
  pages        = {14431--14440},
  publisher    = {Computer Vision Foundation / {IEEE}},
  year         = {2020},
}

@inproceedings{DBLP:conf/ijcai/ZhangHWZZWZ024,
  author       = {Hangtao Zhang and
                  Shengshan Hu and
                  Yichen Wang and
                  Leo Yu Zhang and
                  Ziqi Zhou and
                  Xianlong Wang and
                  Yanjun Zhang and
                  Chao Chen},
  title        = {Detector Collapse: Backdooring Object Detection to Catastrophic Overload
                  or Blindness in the Physical World},
  booktitle    = {Proceedings of the Thirty-Third International Joint Conference on
                  Artificial Intelligence, {IJCAI} 2024, Jeju, South Korea, August 3-9,
                  2024},
  pages        = {1670--1678},
  publisher    = {ijcai.org},
  year         = {2024},
}

@article{DBLP:journals/corr/abs-2411-14243,
  author       = {Jialin Lu and
                  Junjie Shan and
                  Ziqi Zhao and
                  Ka{-}Ho Chow},
  title        = {AnywhereDoor: Multi-Target Backdoor Attacks on Object Detection},
  journal      = {CoRR},
  volume       = {abs/2411.14243},
  year         = {2024},
}

@article{DBLP:journals/corr/abs-2504-16907,
  author       = {Ruotong Wang and
                  Mingli Zhu and
                  Jiarong Ou and
                  Rui Chen and
                  Xin Tao and
                  Pengfei Wan and
                  Baoyuan Wu},
  title        = {BadVideo: Stealthy Backdoor Attack against Text-to-Video Generation},
  journal      = {CoRR},
  volume       = {abs/2504.16907},
  year         = {2025},
}

@inproceedings{DBLP:conf/iccv/KirillovMRMRGXW23,
  author       = {Alexander Kirillov and
                  Eric Mintun and
                  Nikhila Ravi and
                  Hanzi Mao and
                  Chlo{\'{e}} Rolland and
                  Laura Gustafson and
                  Tete Xiao and
                  Spencer Whitehead and
                  Alexander C. Berg and
                  Wan{-}Yen Lo and
                  Piotr Doll{\'{a}}r and
                  Ross B. Girshick},
  title        = {Segment Anything},
  booktitle    = {{IEEE/CVF} International Conference on Computer Vision, {ICCV} 2023,
                  Paris, France, October 1-6, 2023},
  pages        = {3992--4003},
  publisher    = {{IEEE}},
  year         = {2023},
}

@article{DBLP:journals/corr/abs-2503-12781,
  author       = {Zhang Jiaxing and
                  Tang Hao},
  title        = {{SAM2} for Image and Video Segmentation: {A} Comprehensive Survey},
  journal      = {CoRR},
  volume       = {abs/2503.12781},
  year         = {2025},
}

@inproceedings{Zhang2024PathSAM2TS,
  title={Path-SAM2: Transfer SAM2 for digital pathology semantic segmentation},
  author={Mingya Zhang and Liang Wang and Zhihao Chen and Yiyuan Ge and Xianping Tao},
  year={2024},
  url={https://api.semanticscholar.org/CorpusID:271745331}
}

@article{DBLP:journals/corr/abs-2408-00874,
  author       = {Jiayuan Zhu and
                  Yunli Qi and
                  Junde Wu},
  title        = {Medical {SAM} 2: Segment medical images as video via Segment Anything
                  Model 2},
  journal      = {CoRR},
  volume       = {abs/2408.00874},
  year         = {2024},
}

@article{DBLP:journals/corr/abs-2408-03286,
  author       = {Zhiling Yan and
                  Weixiang Sun and
                  Rong Zhou and
                  Zhengqing Yuan and
                  Kai Zhang and
                  Yiwei Li and
                  Tianming Liu and
                  Quanzheng Li and
                  Xiang Li and
                  Lifang He and
                  Lichao Sun},
  title        = {Biomedical {SAM} 2: Segment Anything in Biomedical Images and Videos},
  journal      = {CoRR},
  volume       = {abs/2408.03286},
  year         = {2024},
}

@article{DBLP:journals/corr/abs-1712-05526,
  author       = {Xinyun Chen and
                  Chang Liu and
                  Bo Li and
                  Kimberly Lu and
                  Dawn Song},
  title        = {Targeted Backdoor Attacks on Deep Learning Systems Using Data Poisoning},
  journal      = {CoRR},
  volume       = {abs/1712.05526},
  year         = {2017},
}

@article{DBLP:journals/corr/abs-1809-03327,
  author       = {Ning Xu and
                  Linjie Yang and
                  Yuchen Fan and
                  Dingcheng Yue and
                  Yuchen Liang and
                  Jianchao Yang and
                  Thomas S. Huang},
  title        = {YouTube-VOS: {A} Large-Scale Video Object Segmentation Benchmark},
  journal      = {CoRR},
  volume       = {abs/1809.03327},
  year         = {2018},
}

@inproceedings{DBLP:conf/eccv/MeiZYYQCK22,
  author       = {Jieru Mei and
                  Alex Zihao Zhu and
                  Xinchen Yan and
                  Hang Yan and
                  Siyuan Qiao and
                  Liang{-}Chieh Chen and
                  Henrik Kretzschmar},
  editor       = {Shai Avidan and
                  Gabriel J. Brostow and
                  Moustapha Ciss{\'{e}} and
                  Giovanni Maria Farinella and
                  Tal Hassner},
  title        = {Waymo Open Dataset: Panoramic Video Panoptic Segmentation},
  booktitle    = {Computer Vision - {ECCV} 2022 - 17th European Conference, Tel Aviv,
                  Israel, October 23-27, 2022, Proceedings, Part {XXIX}},
  series       = {Lecture Notes in Computer Science},
  volume       = {13689},
  pages        = {53--72},
  publisher    = {Springer},
  year         = {2022},
}

@misc{cvpr2018_autonomous_driving,
  title        = {CVPR 2018 WAD Video Segmentation Challenge (Autonomous Driving)},
  howpublished = {Kaggle competition},
  year         = {2018},
}

@inproceedings{DBLP:conf/iccv/LiKHTR13,
  author       = {Fuxin Li and
                  Taeyoung Kim and
                  Ahmad Humayun and
                  David Tsai and
                  James M. Rehg},
  title        = {Video Segmentation by Tracking Many Figure-Ground Segments},
  booktitle    = {{IEEE} International Conference on Computer Vision, {ICCV} 2013, Sydney,
                  Australia, December 1-8, 2013},
  pages        = {2192--2199},
  publisher    = {{IEEE} Computer Society},
  year         = {2013},
  url          = {https://doi.org/10.1109/ICCV.2013.273},
  doi          = {10.1109/ICCV.2013.273},
  bibsource    = {dblp computer science bibliography, https://dblp.org}
}

@inproceedings{DBLP:conf/iclr/NguyenT21,
  author       = {Tuan Anh Nguyen and
                  Anh Tuan Tran},
  title        = {WaNet - Imperceptible Warping-based Backdoor Attack},
  booktitle    = {9th International Conference on Learning Representations, {ICLR} 2021,
                  Virtual Event, Austria, May 3-7, 2021},
  publisher    = {OpenReview.net},
  year         = {2021},
}

@inproceedings{DBLP:conf/sp/WangYSLVZZ19,
  author       = {Bolun Wang and
                  Yuanshun Yao and
                  Shawn Shan and
                  Huiying Li and
                  Bimal Viswanath and
                  Haitao Zheng and
                  Ben Y. Zhao},
  title        = {Neural Cleanse: Identifying and Mitigating Backdoor Attacks in Neural
                  Networks},
  booktitle    = {2019 {IEEE} Symposium on Security and Privacy, {SP} 2019, San Francisco,
                  CA, USA, May 19-23, 2019},
  pages        = {707--723},
  publisher    = {{IEEE}},
  year         = {2019},
}

@inproceedings{DBLP:conf/acsac/GaoXW0RN19,
  author       = {Yansong Gao and
                  Chang Xu and
                  Derui Wang and
                  Shiping Chen and
                  Damith Chinthana Ranasinghe and
                  Surya Nepal},
  editor       = {David M. Balenson},
  title        = {{STRIP:} a defence against trojan attacks on deep neural networks},
  booktitle    = {Proceedings of the 35th Annual Computer Security Applications Conference,
                  {ACSAC} 2019, San Juan, PR, USA, December 09-13, 2019},
  pages        = {113--125},
  publisher    = {{ACM}},
  year         = {2019},
}

@inproceedings{DBLP:conf/iclr/Huang0WQ022,
  author       = {Kunzhe Huang and
                  Yiming Li and
                  Baoyuan Wu and
                  Zhan Qin and
                  Kui Ren},
  title        = {Backdoor Defense via Decoupling the Training Process},
  booktitle    = {The Tenth International Conference on Learning Representations, {ICLR}
                  2022, Virtual Event, April 25-29, 2022},
  publisher    = {OpenReview.net},
  year         = {2022},
}

@inproceedings{DBLP:conf/raid/0017DG18,
  author       = {Kang Liu and
                  Brendan Dolan{-}Gavitt and
                  Siddharth Garg},
  editor       = {Michael D. Bailey and
                  Thorsten Holz and
                  Manolis Stamatogiannakis and
                  Sotiris Ioannidis},
  title        = {Fine-Pruning: Defending Against Backdooring Attacks on Deep Neural
                  Networks},
  booktitle    = {Research in Attacks, Intrusions, and Defenses - 21st International
                  Symposium, {RAID} 2018, Heraklion, Crete, Greece, September 10-12,
                  2018, Proceedings},
  series       = {Lecture Notes in Computer Science},
  volume       = {11050},
  pages        = {273--294},
  publisher    = {Springer},
  year         = {2018},
}

@misc{sam2,
  author       = {Meta AI Research},
  title        = {SAM2: Segment Anything Model 2},
  year         = {2024},
  howpublished = {\url{https://github.com/facebookresearch/sam2}},
  note         = {Accessed: 2025-08-05}
}

@article{DBLP:journals/corr/abs-2411-19530,
  author       = {Yule Liu and
                  Zhen Sun and
                  Xinlei He and
                  Xinyi Huang},
  title        = {Quantized Delta Weight Is Safety Keeper},
  journal      = {CoRR},
  volume       = {abs/2411.19530},
  year         = {2024},
}

@article{DBLP:journals/corr/abs-1710-00942,
  author       = {Yuntao Liu and
                  Yang Xie and
                  Ankur Srivastava},
  title        = {Neural Trojans},
  journal      = {CoRR},
  volume       = {abs/1710.00942},
  year         = {2017},
}

@inproceedings{DBLP:conf/cvpr/WengerPBY0Z21,
  author       = {Emily Wenger and
                  Josephine Passananti and
                  Arjun Nitin Bhagoji and
                  Yuanshun Yao and
                  Haitao Zheng and
                  Ben Y. Zhao},
  title        = {Backdoor Attacks Against Deep Learning Systems in the Physical World},
  booktitle    = {{IEEE} Conference on Computer Vision and Pattern Recognition, {CVPR}
                  2021, virtual, June 19-25, 2021},
  pages        = {6206--6215},
  publisher    = {Computer Vision Foundation / {IEEE}},
  year         = {2021},
  url          = {https://openaccess.thecvf.com/content/CVPR2021/html/Wenger\_Backdoor\_Attacks\_Against\_Deep\_Learning\_Systems\_in\_the\_Physical\_World\_CVPR\_2021\_paper.html},
  doi          = {10.1109/CVPR46437.2021.00614},
  bibsource    = {dblp computer science bibliography, https://dblp.org}
}

@inproceedings{DBLP:conf/mm/HanXZYLZ22,
  author       = {Xingshuo Han and
                  Guowen Xu and
                  Yuan Zhou and
                  Xuehuan Yang and
                  Jiwei Li and
                  Tianwei Zhang},
  editor       = {Jo{\~{a}}o Magalh{\~{a}}es and
                  Alberto Del Bimbo and
                  Shin'ichi Satoh and
                  Nicu Sebe and
                  Xavier Alameda{-}Pineda and
                  Qin Jin and
                  Vincent Oria and
                  Laura Toni},
  title        = {Physical Backdoor Attacks to Lane Detection Systems in Autonomous
                  Driving},
  booktitle    = {{MM} '22: The 30th {ACM} International Conference on Multimedia, Lisboa,
                  Portugal, October 10 - 14, 2022},
  pages        = {2957--2968},
  publisher    = {{ACM}},
  year         = {2022},
}

@article{DBLP:journals/corr/abs-2410-16268,
  author       = {Shuangrui Ding and
                  Rui Qian and
                  Xiaoyi Dong and
                  Pan Zhang and
                  Yuhang Zang and
                  Yuhang Cao and
                  Yuwei Guo and
                  Dahua Lin and
                  Jiaqi Wang},
  title        = {SAM2Long: Enhancing {SAM} 2 for Long Video Segmentation with a Training-Free
                  Memory Tree},
  journal      = {CoRR},
  volume       = {abs/2410.16268},
  year         = {2024},
  url          = {https://doi.org/10.48550/arXiv.2410.16268},
  doi          = {10.48550/ARXIV.2410.16268},
  eprinttype    = {arXiv},
  eprint       = {2410.16268},
  bibsource    = {dblp computer science bibliography, https://dblp.org}
}

@inproceedings{DBLP:conf/cvpr/ZhouZXSXWK0LCS25,
  author       = {Chong Zhou and
                  Chenchen Zhu and
                  Yunyang Xiong and
                  Saksham Suri and
                  Fanyi Xiao and
                  Lemeng Wu and
                  Raghuraman Krishnamoorthi and
                  Bo Dai and
                  Chen Change Loy and
                  Vikas Chandra and
                  Bilge Soran},
  title        = {EdgeTAM: On-Device Track Anything Model},
  booktitle    = {{IEEE/CVF} Conference on Computer Vision and Pattern Recognition,
                  {CVPR} 2025, Nashville, TN, USA, June 11-15, 2025},
  pages        = {13832--13842},
  publisher    = {Computer Vision Foundation / {IEEE}},
  year         = {2025},
  url          = {https://openaccess.thecvf.com/content/CVPR2025/html/Zhou\_EdgeTAM\_On-Device\_Track\_Anything\_Model\_CVPR\_2025\_paper.html},
  doi          = {10.1109/CVPR52734.2025.01291},
  bibsource    = {dblp computer science bibliography, https://dblp.org}
}

@inproceedings{DBLP:conf/cvpr/FengMZZXT22,
  author       = {Yu Feng and
                  Benteng Ma and
                  Jing Zhang and
                  Shanshan Zhao and
                  Yong Xia and
                  Dacheng Tao},
  title        = {{FIBA:} Frequency-Injection based Backdoor Attack in Medical Image
                  Analysis},
  booktitle    = {{IEEE/CVF} Conference on Computer Vision and Pattern Recognition,
                  {CVPR} 2022, New Orleans, LA, USA, June 18-24, 2022},
  pages        = {20844--20853},
  publisher    = {{IEEE}},
  year         = {2022},
  url          = {https://doi.org/10.1109/CVPR52688.2022.02021},
  doi          = {10.1109/CVPR52688.2022.02021},
  bibsource    = {dblp computer science bibliography, https://dblp.org}
}

@article{DBLP:journals/corr/Pont-TusetPCASG17,
  author       = {Jordi Pont{-}Tuset and
                  Federico Perazzi and
                  Sergi Caelles and
                  Pablo Arbel{\'{a}}ez and
                  Alexander Sorkine{-}Hornung and
                  Luc Van Gool},
  title        = {The 2017 {DAVIS} Challenge on Video Object Segmentation},
  journal      = {CoRR},
  volume       = {abs/1704.00675},
  year         = {2017},
  url          = {http://arxiv.org/abs/1704.00675},
  eprinttype    = {arXiv},
  eprint       = {1704.00675},
  bibsource    = {dblp computer science bibliography, https://dblp.org}
}

@inproceedings{DBLP:conf/acl/AbnarZ20,
  author       = {Samira Abnar and
                  Willem H. Zuidema},
  editor       = {Dan Jurafsky and
                  Joyce Chai and
                  Natalie Schluter and
                  Joel R. Tetreault},
  title        = {Quantifying Attention Flow in Transformers},
  booktitle    = {Proceedings of the 58th Annual Meeting of the Association for Computational
                  Linguistics, {ACL} 2020, Online, July 5-10, 2020},
  pages        = {4190--4197},
  publisher    = {Association for Computational Linguistics},
  year         = {2020},
  url          = {https://doi.org/10.18653/v1/2020.acl-main.385},
  doi          = {10.18653/V1/2020.ACL-MAIN.385},
  bibsource    = {dblp computer science bibliography, https://dblp.org}
}

@inproceedings{DBLP:conf/nips/Tran0M18,
  author       = {Brandon Tran and
                  Jerry Li and
                  Aleksander Madry},
  editor       = {Samy Bengio and
                  Hanna M. Wallach and
                  Hugo Larochelle and
                  Kristen Grauman and
                  Nicol{\`{o}} Cesa{-}Bianchi and
                  Roman Garnett},
  title        = {Spectral Signatures in Backdoor Attacks},
  booktitle    = {Advances in Neural Information Processing Systems 31: Annual Conference
                  on Neural Information Processing Systems 2018, NeurIPS 2018, December
                  3-8, 2018, Montr{\'{e}}al, Canada},
  pages        = {8011--8021},
  year         = {2018},
  url          = {https://proceedings.neurips.cc/paper/2018/hash/280cf18baf4311c92aa5a042336587d3-Abstract.html},
  bibsource    = {dblp computer science bibliography, https://dblp.org}
}

@inproceedings{DBLP:conf/nips/YuK0LHF20,
  author       = {Tianhe Yu and
                  Saurabh Kumar and
                  Abhishek Gupta and
                  Sergey Levine and
                  Karol Hausman and
                  Chelsea Finn},
  editor       = {Hugo Larochelle and
                  Marc'Aurelio Ranzato and
                  Raia Hadsell and
                  Maria{-}Florina Balcan and
                  Hsuan{-}Tien Lin},
  title        = {Gradient Surgery for Multi-Task Learning},
  booktitle    = {Advances in Neural Information Processing Systems 33: Annual Conference
                  on Neural Information Processing Systems 2020, NeurIPS 2020, December
                  6-12, 2020, virtual},
  year         = {2020},
  url          = {https://proceedings.neurips.cc/paper/2020/hash/3fe78a8acf5fda99de95303940a2420c-Abstract.html},
  bibsource    = {dblp computer science bibliography, https://dblp.org}
}

@inproceedings{DBLP:conf/acl/Gu0LL0W23,
  author       = {Naibin Gu and
                  Peng Fu and
                  Xiyu Liu and
                  Zhengxiao Liu and
                  Zheng Lin and
                  Weiping Wang},
  editor       = {Anna Rogers and
                  Jordan L. Boyd{-}Graber and
                  Naoaki Okazaki},
  title        = {A Gradient Control Method for Backdoor Attacks on Parameter-Efficient
                  Tuning},
  booktitle    = {Proceedings of the 61st Annual Meeting of the Association for Computational
                  Linguistics (Volume 1: Long Papers), {ACL} 2023, Toronto, Canada,
                  July 9-14, 2023},
  pages        = {3508--3520},
  publisher    = {Association for Computational Linguistics},
  year         = {2023},
  url          = {https://doi.org/10.18653/v1/2023.acl-long.194},
  doi          = {10.18653/V1/2023.ACL-LONG.194},
  bibsource    = {dblp computer science bibliography, https://dblp.org}
}

@inproceedings{DBLP:conf/nips/ZhaoMCSWAYZLC24,
  author       = {Haozhe Zhao and
                  Xiaojian (Shawn) Ma and
                  Liang Chen and
                  Shuzheng Si and
                  Rujie Wu and
                  Kaikai An and
                  Peiyu Yu and
                  Minjia Zhang and
                  Qing Li and
                  Baobao Chang},
  editor       = {Amir Globersons and
                  Lester Mackey and
                  Danielle Belgrave and
                  Angela Fan and
                  Ulrich Paquet and
                  Jakub M. Tomczak and
                  Cheng Zhang},
  title        = {UltraEdit: Instruction-based Fine-Grained Image Editing at Scale},
  booktitle    = {Advances in Neural Information Processing Systems 38: Annual Conference
                  on Neural Information Processing Systems 2024, NeurIPS 2024, Vancouver,
                  BC, Canada, December 10 - 15, 2024},
  year         = {2024},
  url          = {http://papers.nips.cc/paper\_files/paper/2024/hash/05a30a0fc9e6bacdd3abd4ca8508a9e6-Abstract-Datasets\_and\_Benchmarks\_Track.html},
  bibsource    = {dblp computer science bibliography, https://dblp.org}
}

@inproceedings{guo2026sixdattack,
  author    = {Jihui Guo and
               Zongmin Zhang and
               Zhen Sun and
               Yuhao Yang and
               Jinlin Wu and
               Fu Zhang and
               Xinlei He},
  title     = {6DAttack: Backdoor Attacks in the 6DoF Pose Estimation},
  booktitle = {AAAI Conference on Artificial Intelligence (AAAI)},
  year      = {2026}
}

@InProceedings{Amula_2025_ICCV,
    author    = {Amula, Venkat Adithya and Samavedam, Sunayana and Saini, Saurabh and Gupta, Avani and Narayanan, P J},
    title     = {Prototype Guided Backdoor Defense via Activation Space Manipulation},
    booktitle = {Proceedings of the IEEE/CVF International Conference on Computer Vision (ICCV)},
    month     = {October},
    year      = {2025},
    pages     = {2195-2205}
}

@inproceedings{DBLP:conf/sp/JiaLG22,
  author       = {Jinyuan Jia and
                  Yupei Liu and
                  Neil Zhenqiang Gong},
  title        = {BadEncoder: Backdoor Attacks to Pre-trained Encoders in Self-Supervised
                  Learning},
  booktitle    = {43rd {IEEE} Symposium on Security and Privacy, {SP} 2022, San Francisco,
                  CA, USA, May 22-26, 2022},
  pages        = {2043--2059},
  publisher    = {{IEEE}},
  year         = {2022},
  url          = {https://doi.org/10.1109/SP46214.2022.9833644},
  doi          = {10.1109/SP46214.2022.9833644},
  bibsource    = {dblp computer science bibliography, https://dblp.org}
}

@article{DBLP:journals/corr/abs-2505-16640,
  author       = {Xueyang Zhou and
                  Guiyao Tie and
                  Guowen Zhang and
                  Hechang Wang and
                  Pan Zhou and
                  Lichao Sun},
  title        = {BadVLA: Towards Backdoor Attacks on Vision-Language-Action Models
                  via Objective-Decoupled Optimization},
  journal      = {CoRR},
  volume       = {abs/2505.16640},
  year         = {2025},
  url          = {https://doi.org/10.48550/arXiv.2505.16640},
  doi          = {10.48550/ARXIV.2505.16640},
  eprinttype   = {arXiv},
  eprint       = {2505.16640},
  bibsource    = {dblp computer science bibliography, https://dblp.org}
}

\clearpage
\appendix
\section{Ethics Considerations}
This study aims to reveal and investigate overlooked security risks in VSFMs, particularly in the context of backdoor attacks. 
Although the proposed framework, \Method, demonstrates an effective method for implanting backdoors into VSFMs, our intention is not to promote or disseminate malicious attack techniques. 
Instead, our goal is to raise awareness within the research community about the unique vulnerabilities of VSFMs and to lay the groundwork for building more robust, interpretable, and secure video segmentation systems.

All experiments are conducted responsibly in controlled environments using publicly available datasets such as LVOS and DAVIS. 
We do not release any poisoned models or datasets with triggers. 
No human data or privacy-sensitive information is involved in this study.
In future work, we plan to explore effective defense mechanisms and detection techniques for VSFMs, aiming to enhance the reliability and security of these powerful models before their deployment in safety-critical applications, such as autonomous driving, medical imaging, and robotics.

\section{Additional Figures and Tables}
\begin{center}
\small
\setlength{\tabcolsep}{4pt}
\renewcommand{\arraystretch}{1.12}
\captionof{table}{Robustness of \Method with BadNet at a 5\% poisoning rate under physical-world trigger transformations on DAVIS.}
\label{tab:physical_transform}
\resizebox{\linewidth}{!}{
\begin{tabular}{lccc}
\toprule
Transformation & Point ASR~(\%) & Box ASR~(\%) & Mask ASR~(\%) \\
\midrule
Original & 91.8 & 93.1 & 42.7 \\
Bright & 91.7 & 93.8 & 43.3 \\
Dark & 91.6 & 91.9 & 35.9 \\
Color cast & 91.9 & 92.4 & 44.0 \\
Color shift & 92.0 & 93.3 & 42.1 \\
JPEG Q=10 & 98.0 & 96.8 & 76.1 \\
JPEG Q=5 & 93.0 & 91.3 & 73.5 \\
Double JPEG & 93.5 & 96.8 & 63.8 \\
Global noise & 92.6 & 92.8 & 55.0 \\
Noisy trigger & 92.6 & 91.8 & 44.7 \\
Gaussian blur & 89.9 & 89.8 & 38.3 \\
Global motion blur & 85.7 & 77.9 & 41.4 \\
Local motion blur & 73.8 & 63.0 & 27.0 \\
Position jitter & 90.0 & 83.2 & 40.5 \\
Rotation $15^\circ$ & 90.5 & 90.9 & 41.8 \\
Rotation $30^\circ$ & 89.1 & 84.2 & 41.0 \\
Rotation $45^\circ$ & 82.5 & 77.5 & 41.6 \\
Perspective & 92.1 & 89.8 & 45.3 \\
50\% occlusion & 90.7 & 88.4 & 42.5 \\
75\% occlusion & 68.1 & 66.0 & 35.7 \\
$2\times$ scale & 89.6 & 89.6 & 39.8 \\
$0.5\times$ scale & 76.4 & 75.9 & 36.3 \\
Low resolution & 92.8 & 90.9 & 39.3 \\
\bottomrule
\end{tabular}}
\end{center}

\begin{figure*}[t]
  \centering
  \includegraphics[width=\textwidth]{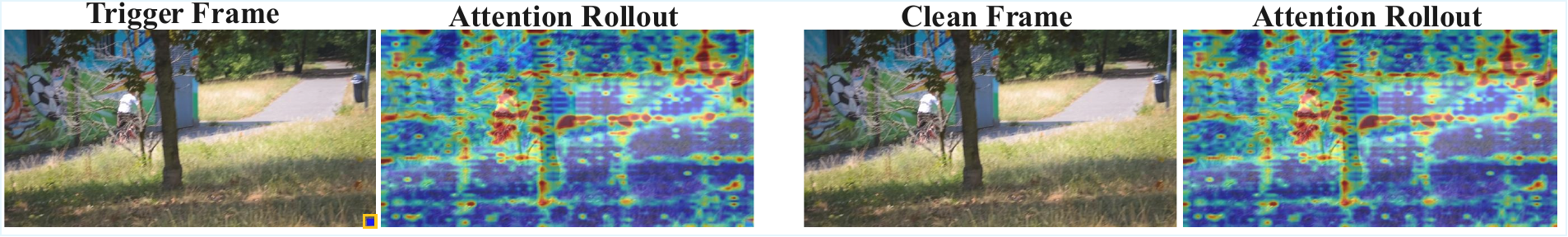}
  \caption{Visualization results for attention maps of the SAM2 image encoder trained with BadNet on DAVIS at a 5\% poisoning rate.}
  \label{fig:badnet_attention}
\end{figure*}

\end{document}